%% file: main.tex
\documentclass[10pt,logo,copyright]{nvidiatechreport}
\input{packages}
\input{common}

\input{merged_macros}

\usepackage{tikz}
\usetikzlibrary{arrows.meta,positioning}
\usepackage{pifont}
\usepackage{diagbox}
\usepackage{colortbl}
\usepackage{graphicx}
\usepackage{wrapfig}
\usepackage{mathtools}
\usepackage{placeins}

\usepackage[nameinlink]{cleveref}
\crefname{equation}{Eq.}{Eqs.}
\crefname{figure}{Fig.}{Figs.}
\crefname{section}{Sec.}{Sec.}
\crefname{appendix}{App.}{App.}
\crefname{table}{Tab.}{Tabs.}
\crefname{algorithm}{Algo}{Algo}
\crefname{thm}{Thm}{Thm}
\Crefname{thm}{Thm}{Thm}
\crefname{prop}{Prop}{Prop}

\definecolor{darkred}{rgb}{0.7, 0.0, 0.0}

\newcommand{\method}{\textsc{SR-ReaL}}

\title{Reinforcing Dual-Path Reasoning in Spatial Vision Language Models}

\author{%
\textbf{Yatai Ji}$^{1,2,\ast}$\hspace{0.45em}
\textbf{An-Chieh Cheng}$^{2,3}$\hspace{0.45em}
\textbf{Yang Fu}$^{2,3}$\hspace{0.45em}
\textbf{Yukang Chen}$^{2}$\hspace{0.45em}
\textbf{Han Zhang}$^{2}$\hspace{0.45em}
\textbf{Zhaojing Yang}$^{3}$\hspace{0.45em}
\textbf{Wei Huang}$^{1,2}$\hspace{0.45em}
\textbf{Ka Chun Cheung}$^{2}$\hspace{0.45em}
\textbf{Song Han}$^{2}$\hspace{0.45em}
\textbf{Vidya Nariyambut Murali}$^{2}$\hspace{0.45em}
\textbf{Pavlo Molchanov}$^{2}$\hspace{0.45em}
\textbf{Simon See}$^{2}$\hspace{0.45em}
\textbf{Jan Kautz}$^{2}$\hspace{0.45em}
\textbf{Hongxu Yin}$^{2}$\hspace{0.45em}
\textbf{Ping Luo}$^{1,\dagger}$\hspace{0.45em}
\textbf{Sifei Liu}$^{2,\dagger}$%
  
\par\vspace{7pt}
{$^1$\,The University of Hong Kong \quad $^2$\,NVIDIA \quad $^3$\,University of California, San Diego}%
}%

\input{sec/0_abstract_sf}

\defcitealias{hatamizadeh2024mambavision}{Hatamizadeh et al. (2024)}

\begin{document}
\maketitle

\begingroup
\renewcommand{\thefootnote}{\fnsymbol{footnote}}
\footnotetext[1]{Work done during an internship at NVIDIA.}
\footnotetext[2]{Corresponding author.}
\endgroup

\vspace{-1.0em}
\noindent
\small
\textbf{Project page:} \href{https://sr-real.github.io/}{https://sr-real.github.io/}
\par\vspace{1.3em}

\abscontent

\newpage
\tableofcontents
\newpage

\input{sec/1_intro_sf2}
\input{sec/2_relatedwork}

\input{sec/3_method}

\input{sec/4_exp}
\input{sec/5_conl}

\clearpage
\appendix
\renewcommand{\thefigure}{S\arabic{figure}}
\setcounter{figure}{0}
\renewcommand{\thetable}{S\arabic{table}}
\setcounter{table}{0}

\input{sec/X_suppl}

\clearpage
\begingroup
\raggedright
\sloppy
\makeatletter
\renewcommand{\bibsection}{%
  \par\noindent{\headingfont\refname}\par\vspace{5pt}%
}
\setlength{\bibhang}{0pt}
\renewcommand\@bibsetup[1]{%
  \setlength{\leftmargin}{0pt}%
  \setlength{\itemindent}{0pt}%
  \setlength{\labelwidth}{0pt}%
  \setlength{\labelsep}{0pt}%
  \setlength{\listparindent}{0pt}%
  \setlength{\itemsep}{\bibsep}%
}
\makeatother
\setlength{\emergencystretch}{3em}
\Urlmuskip=0mu plus 1mu\relax
\bibliographystyle{plainnat}
\bibliography{main}
\endgroup

\end{document}

%% file: packages.tex
\usepackage[authoryear,round]{natbib}

\usepackage[utf8]{inputenc} 
\usepackage[T1]{fontenc}    

\usepackage{parskip}        
\usepackage{url}            
\usepackage{booktabs}       
\usepackage{amsfonts}       
\usepackage{nicefrac}       
\usepackage{microtype}      
\usepackage{xcolor}         
\usepackage[dvipsnames]{xcolor} 
\usepackage{graphicx}
\usepackage{animate}        
\usepackage{subcaption}
\usepackage{tabularx}
\usepackage{makecell}
\usepackage{adjustbox}
\usepackage{setspace}
\newcolumntype{M}[1]{>{\centering\arraybackslash}m{#1}}
\usepackage{float}
\usepackage{tikz}
\usetikzlibrary{positioning,shapes,shapes.geometric,arrows, decorations.pathreplacing, backgrounds}
\usepackage{amsmath,amsfonts,bm, bbm}
\usepackage{multirow}
\usepackage{comment}
\usepackage{gensymb}
\usepackage{lipsum}
\usetikzlibrary{arrows.meta, positioning, fit}
\usepackage[para]{threeparttable}
\usetikzlibrary{tikzmark}

\usepackage[most]{tcolorbox}
\usepackage{fancyvrb}
\usepackage{fvextra}
\usepackage{dashrule}

\usepackage{multicol}

\makeatletter
\renewcommand{\paragraph}{%
  \@startsection{paragraph}{4}{\z@}%
    {0.5ex \@plus .2ex \@minus .2ex}
    {-0.4em}
    {\normalfont\normalsize\bfseries}%
}
\makeatother

%% file: common.tex
\def\eqref#1{equation~\ref{#1}}

\def\1{\bm{1}}

\DeclareMathAlphabet{\mathsfit}{\encodingdefault}{\sfdefault}{m}{sl}
\SetMathAlphabet{\mathsfit}{bold}{\encodingdefault}{\sfdefault}{bx}{n}

\makeatletter
\let\save@mathaccent\mathaccent
\newcommand*\if@single[3]{%
  \setbox0\hbox{${\mathaccent"0362{#1}}^H$}%
  \setbox2\hbox{${\mathaccent"0362{\kern0pt#1}}^H$}%
  \ifdim\ht0=\ht2 #3\else #2\fi
  }
\newcommand*\rel@kern[1]{\kern#1\dimexpr\macc@kerna}
\newcommand*\widebar[1]{\@ifnextchar^{{\wide@bar{#1}{0}}}{\wide@bar{#1}{1}}}
\newcommand*\wide@bar[2]{\if@single{#1}{\wide@bar@{#1}{#2}{1}}{\wide@bar@{#1}{#2}{2}}}
\newcommand*\wide@bar@[3]{%
  \begingroup
  \def\mathaccent##1##2{%
    \let\mathaccent\save@mathaccent
    \if#32 \let\macc@nucleus\first@char \fi
    \setbox\z@\hbox{$\macc@style{\macc@nucleus}_{}$}%
    \setbox\tw@\hbox{$\macc@style{\macc@nucleus}{}_{}$}%
    \dimen@\wd\tw@
    \advance\dimen@-\wd\z@
    \divide\dimen@ 3
    \@tempdima\wd\tw@
    \advance\@tempdima-\scriptspace
    \divide\@tempdima 10
    \advance\dimen@-\@tempdima
    \ifdim\dimen@>\z@ \dimen@0pt\fi
    \rel@kern{0.6}\kern-\dimen@
    \if#31
      \overline{\rel@kern{-0.6}\kern\dimen@\macc@nucleus\rel@kern{0.4}\kern\dimen@}%
      \advance\dimen@0.4\dimexpr\macc@kerna
      \let\final@kern#2%
      \ifdim\dimen@<\z@ \let\final@kern1\fi
      \if\final@kern1 \kern-\dimen@\fi
    \else
      \overline{\rel@kern{-0.6}\kern\dimen@#1}%
    \fi
  }%
  \macc@depth\@ne
  \let\math@bgroup\@empty \let\math@egroup\macc@set@skewchar
  \mathsurround\z@ \frozen@everymath{\mathgroup\macc@group\relax}%
  \macc@set@skewchar\relax
  \let\mathaccentV\macc@nested@a
  \if#31
    \macc@nested@a\relax111{#1}%
  \else
    \def\gobble@till@marker##1\endmarker{}%
    \futurelet\first@char\gobble@till@marker#1\endmarker
    \ifcat\noexpand\first@char A\else
      \def\first@char{}%
    \fi
    \macc@nested@a\relax111{\first@char}%
  \fi
  \endgroup
}
\makeatother


%% file: merged_macros.tex
\newcommand{\whj}[1]{}
\newcommand{\SL}[1]{}
\newcommand{\wonmin}[1]{}
\newcommand{\jinwei}[1]{}
\newcommand{\kh}[1]{}
\newcommand{\ye}[1]{}
\newcommand{\YT}[1]{}
\newcommand{\CM}[1]{}
\newcommand{\Sifei}[1]{}
\newcommand{\Yitong}[1]{}
\newcommand{\Hongju}[1]{}
\newcommand{\Collin}[1]{}
\newcommand{\Jinwei}[1]{}
\newcommand{\Pavlo}[1]{}
\newcommand{\Tianfan}[1]{}

\definecolor{Green}{HTML}{E5F8F6}
\definecolor{Blue}{HTML}{DAE3F5}
\definecolor{Purple}{HTML}{ECE3EC}
\definecolor{RPurple}{HTML}{FED8B1}
\definecolor{Yellow}{HTML}{fdffb7}
\definecolor{Grey}{rgb}{0.81176471, 0.81176471, 0.81176471}
\definecolor{DGreen}{rgb}{0.15882353, 0.80980392, 0.14705882}
\definecolor{ngreen}{HTML}{76B900}



%% file: sec/0_abstract_sf.tex
\begin{abstract}

Spatial VLMs have made substantial progress in geometric perception, yet complex spatial reasoning—requiring multi-step inference that chains depth cues, distance comparisons, and scene relations—remains challenging. Moreover, different spatial queries demand fundamentally different strategies: some are best resolved through purely linguistic, step-by-step deduction over scene relations, while others benefit from first explicitly grounding objects in 3D space before performing quantitative inference. 
No existing framework simultaneously addresses both strategies within a unified spatial VLM.

We present Dual-Path Spatial Reasoning via Reinforcement Learning for Spatial VLMs (\textbf{\method{}}), a unified framework that equips a spatial VLM with two complementary reasoning paths: Language-Only Reasoning (LOR), which performs step-by-step linguistic deduction, and Detect-Then-Reason (DTR), which first detects 3D geometric cues (e.g., object centers or bounding boxes) via region tokens before explicit geometric inference. \method{} trains both paths through a two-stage pipeline. In the cold-start stage, we construct structured chain-of-thought supervision for LOR and DTR, expose a region-to-3D grounding interface that links visual region tokens to predicted 3D coordinates, and blend 2D/3D grounding data with general-purpose VQA for stable initialization. The subsequent RL stage jointly optimizes both paths using GRPO with accuracy and format rewards; for DTR, a discrete center-based detection reward further refines geometric alignment during reasoning.

Across diverse spatial benchmarks, \method{} significantly outperforms spatial VLM baselines: (i) a single RL-trained checkpoint supports both reasoning paths, with DTR excelling in region-aware tasks through precise 3D localization and LOR enhancing general spatial reasoning; (ii) jointly training both paths fosters mutual reinforcement, with each mode benefiting from the other's supervision; (iii) high-quality, blended cold-start data is crucial for stable RL optimization and cross-domain transfer; and (iv) the model generalizes across datasets and domains without per-task tuning.

\end{abstract}

%% file: sec/1_intro_sf2.tex
\begin{figure*}[ht]
	\centering
	\includegraphics[width=\textwidth]{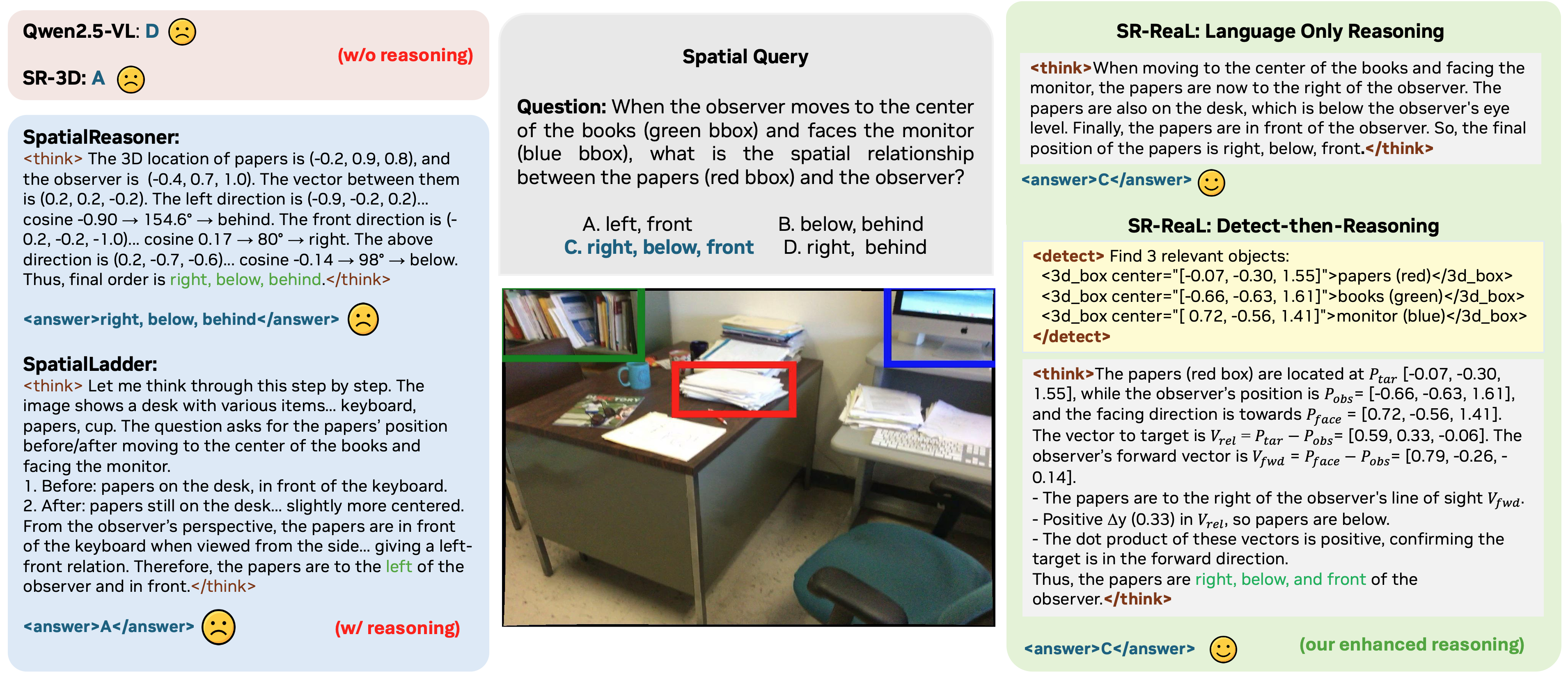}
\caption{A spatial imagination query where \textbf{\method{}} resolves the task under both reasoning paths—Language-Only Reasoning (LOR) and Detect-Then-Reason (DTR) (right-most column). Prior art (left-most column) fail on such examples due to inaccurate or insufficient geometric deduction.}
	\label{fig:teaser}
\end{figure*}

\section{Introduction}
\label{sec:intro}

Large Vision-Language Models (VLMs) have rapidly advanced in interpreting and reasoning over visual content, driven by increasingly capable architectures~\citep{Bai2023QwenVL,Chen2023InternVL,Liu2024LLaVA15,lin2024vila,liu2025nvila,ye2025omnivinci,DBLP:conf/cvpr/JiTJKCZWY023/scl,DBLP:conf/iclr/JiZWSC0YY025/ida-vlm} and proprietary systems~\citep{openai2024gpt4o,google2023gemini}, yet their spatial abilities remain persistently limited—studies reveal weaknesses in understanding 3D layout, depth, occlusion, and viewpoint-dependent relations~\citep{kamath2023whatsup,liu2023visual,wang2024picture,shiri2024empirical}. This has motivated \emph{spatial VLMs}: models that explicitly encode geometric structure and depth cues, enabling more accurate \emph{spatial understanding} of 3D scenes~\citep{Chen2024SpatialVLM,Guo2024RegionGPT,Cheng2024SpatialRGPT,Zhu2024LLaVA3D,Cheng2025SR3D,ma2025spatialllm}. Yet answering challenging spatial questions requires more than recognizing geometric relations—it demands combining multiple cues, applying geometric rules, and reasoning through intermediate steps, a capability we call \emph{spatial reasoning}.

Spatial reasoning goes substantially beyond spatial understanding in both structure and difficulty. While many queries depend on a single local relation, more challenging ones require multi-step inference that chains relations, integrates global context with localized cues, or performs quantitative 3D comparison. Crucially, different spatial queries call for different reasoning strategies: some are best resolved through purely linguistic, step-by-step deduction over scene relations, while queries involving depth, distance, or precise object localization benefit from first grounding objects in 3D space before performing inference. Recent work has applied reinforcement learning (RL) to elicit such multi-step reasoning in language and multimodal models, including R1-style approaches~\citep{deepseekr12025,huang2025visionr1,meng2025mmeureka,yang2025r1onevision,zhang2025r1vl,wang2025vlrethinker} and spatial RL pipelines~\citep{wang2025threedr1,ma2025spatialllm,li2025spatialladder,ma2025spatialreasoner}. However, these systems are built on generic VLMs that lack strong spatial perception and thus cannot leverage the rich geometric structure provided by spatial VLMs. Moreover, no existing method jointly supports both linguistic and geometry-grounded reasoning paths within a single spatial VLM, nor provides the structured supervision to develop them in a unified, mutually reinforcing framework.


We address this problem with \method{}, a unified spatial reasoning model that equips a spatial VLM with two complementary reasoning paths: \emph{Language-Only Reasoning} (LOR), which performs step-by-step linguistic deduction, and \emph{Detect-Then-Reason} (DTR), which injects explicit 3D cues (e.g., centers or oriented boxes) before carrying out quantitative inference. Figure \ref{fig:teaser} shows a spatial query with our LOR and DTR solutions. To enable flexible use of both modes, \method{} trains the model through a two-stage pipeline: a cold-start supervised stage that introduces linguistic and geometry-aware traces while adding the grounding capability missing in the base model, followed by a reinforcement learning stage that refines reasoning using structured feedback. This unified formulation ties the two reasoning paths naturally to both the cold-start and RL procedures.

In the cold-start stage, we construct two structured chain-of-thought (CoT) datasets aligned with LOR and DTR. The LOR dataset provides purely linguistic traces: given a spatial question–answer pair, a large vision-language model is prompted to generate concise step-by-step explanations that derive the answer from scene relations. The DTR dataset augments this format with explicit geometry by including detected or annotated 3D centers or bounding boxes together with reasoning steps that perform quantitative comparison or simple geometric computation.
To enable DTR, we utilize a region-to-3D grounding paradigm with the spatial VLM’s region branch \citep{Cheng2024SpatialRGPT,Cheng2025SR3D}: given a visual region token, the model predicts its 3D center or bounding box, linking region references to geometric quantities through a unified interface. Beyond these two CoT datasets, the cold-start stage also mixes 2D/3D grounding data, region-prompted QA, and general multimodal data so the model acquires grounding and reasoning without losing basic vision–language abilities. Training solely on CoT-style data degrades general multimodal behavior, whereas this blended initialization yields more stable RL optimization and stronger cross-domain transfer.

During RL, \method{} trains both reasoning paths using a DAPO-style Group-Relative Policy Optimization (GRPO) with online filtering. The two paths are prompt-guided: the system prompt explicitly specifies whether to use LOR or DTR, and selecting DTR triggers the model to incorporate extracted region tokens for 3D localization, producing the required 3D predictions before reasoning. The reward combines task accuracy with a format reward and, for DTR, additionally includes a discretized detection reward that facilitates 3D grounding. This setup enables both reasoning modes to be jointly optimized under a unified RL framework while ensuring their compatibility and mutual enhancement.

We evaluate \method{} across a comprehensive suite of spatial benchmarks spanning three categories: (i) region-grounded datasets with localized object references; (ii) global benchmarks requiring whole-scene reasoning; and (iii) out-of-distribution (OOD) sets with shifted imagery or question styles. The results demonstrate that \method{} is highly versatile, effectively supporting both reasoning modes to adapt to diverse scenarios. It consistently outperforms the spatial understanding baseline across multiple benchmarks, with DTR specifically showing superior performance on region-based tasks due to precise 3D localization. Furthermore, we find that jointly training both reasoning modes fosters mutual reinforcement, and improved 3D grounding in DTR directly enhances reasoning accuracy. Our ablation studies also highlight that RL significantly boosts generalization capabilities, while the quality of cold-start data ensures stable RL optimization and robust transfer performance. 

These findings underscore the potential of \method{} in elevating spatial VLMs from perception to reasoning. We demonstrate that with proper grounding and initialization, RL effectively strengthens both linguistic and geometric reasoning skills. In summary, our contributions are: (i) \method{}, a unified dual-path framework that advances spatial reasoning in spatial VLMs via structured RL training; (ii) a structured CoT formulation with two parallel paths—LOR (linguistic) and DTR (geometry-aware) with region-to-3D grounding and a discrete center-based detection reward; and (iii) demonstration of significant performance gains and strong cross-domain generalization, highlighting how region grounding, data quality, and RL supervision jointly determine spatial reasoning behavior.

%% file: sec/2_relatedwork.tex
\section{Related Work}
\label{sec:related work}

\noindent\textbf{Spatial VLMs.}
Early image-based spatial VLMs focus on object-centric properties such as position, size, orientation, and pairwise distance or direction~\citep{kamath2023whatsup,liu2023visual,rajabi2023towards,ranasinghe2024learning,shiri2024empirical,lee2025perspective,wang2024picture,tang2024sparkle,liu2025spatialcot,liu2025ssr,ma2025spatialllm,ma2025spatialreasoner}, with SpatialVLM~\citep{Chen2024SpatialVLM} scaling spatial QA in 2D and SpatialRGPT~\citep{Cheng2024SpatialRGPT} introducing region prompting and depth cues for finer-grained QA. More recent models extend to multi-view or video inputs via 3D positional cues or cross-frame alignment—LLaVA-3D~\citep{Zhu2024LLaVA3D}, Video-3D-LLM~\citep{zheng2025video}, and SR-3D~\citep{Cheng2025SR3D}—enabling multi-view 3D reasoning but assuming known camera poses.

\noindent\textbf{RL for VLM Reasoning.}
Early reasoning advances rely on supervised CoT learning~\citep{muennighoff2025s1,thawakar2025llamavo1}. Building on CoT initialization, RL has emerged as a general mechanism: DeepSeek-R1~\citep{deepseekr12025} shows rule-based RL alone induces long-form reasoning without human-labelled trajectories. In the multimodal setting, Vision-R1 and MM-Eureka~\citep{huang2025visionr1,meng2025mmeureka} adapt R1-style GRPO to mathematical and visual tasks, R1-OneVision~\citep{yang2025r1onevision} converts visual content into structured text before applying RL, and R1-VL/VL-Rethinker~\citep{zhang2025r1vl,wang2025vlrethinker} introduce step-wise reward shaping to improve multimodal reasoning reliability.

\noindent\textbf{Spatial Reasoning VLMs.}
Recent work has focused on equipping vision--language models with stronger spatial reasoning by combining structured spatial supervision, explicit 3D cues, and reinforcement learning. 
3D-R1~\citep{wang2025threedr1} extends R1-style training to multi-view 3D scenes with known camera poses, strengthening 3D spatial reasoning through detailed multi-view textual grounding. 
SpatialReasoner~\citep{ma2025spatialreasoner} and SpatialLLM~\citep{ma2025spatialllm} introduce explicit 3D scene representations that unify perception, spatial computation, and reasoning within a single model, improving their ability to handle diverse 3D spatial questions despite relying solely on single-view RGB inputs. 
Visual Spatial Tuning~\citep{wu2025visualspatialtuning} adopts a progressive pipeline, first improving spatial perception through supervised tasks and then refining spatial CoT via RL. 
SpatialLadder~\citep{li2025spatialladder} employs a three-stage schedule—early 3D localization, structured spatial understanding, and RL-based refinement—to enhance spatial reasoning. 
ViGoRL~\citep{sarch2025vigorl} instead applies multi-turn RL anchored to 2D image coordinates, encouraging the model to iteratively crop, localize, and reason on spatial and visual search tasks. 
In contrast, SR-ReaL conducts RL in a region-based spatial VLM with single-view settings, introducing an internal region-to-3D grounding interface and showing how RL shapes the model's choice between two reasoning paths—LOR and DTR—within one unified architecture.

%% file: sec/3_method.tex
\section{Methods}
\label{sec:method}



\begin{figure*}[t]
	\centering
	\includegraphics[width=\linewidth]{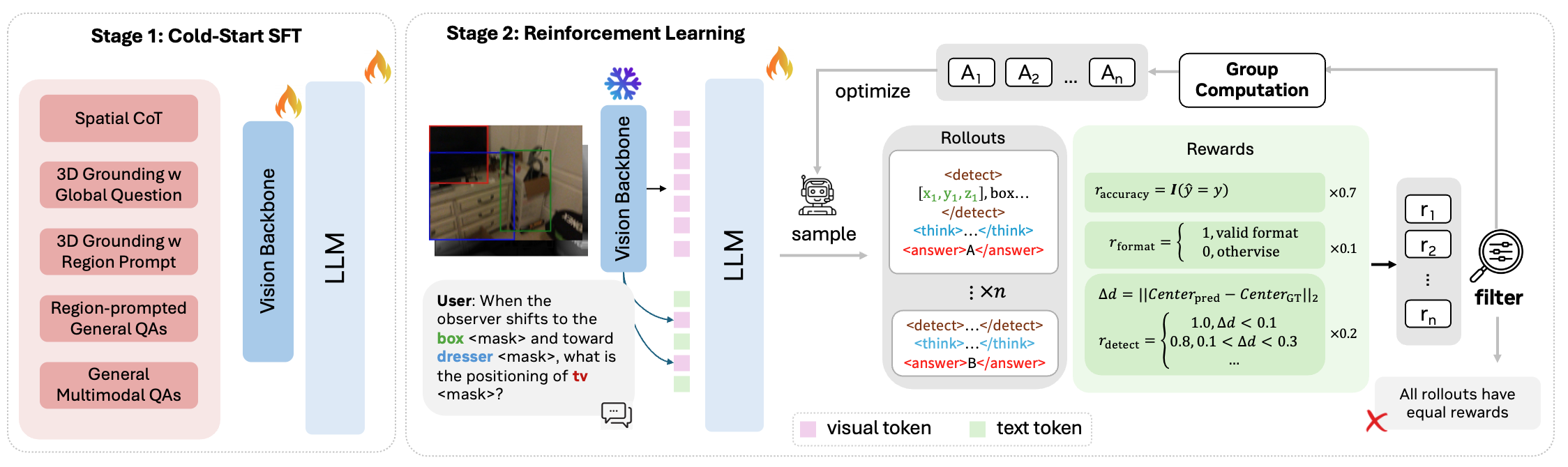}
	\caption{\textbf{Two-stage Training Pipeline:} In Stage 1 (cold-start SFT), we fine-tune the spatial VLM with spatial CoT, 2D/3D grounding (global and region prompts), and region-prompted multimodal QA to initialize spatial reasoning ability. In Stage 2, we apply RL on multiple-choice and filling spatial QA, optimizing grouped rollouts of LOR/DTR trajectories with accuracy, format, and 3D-center detection rewards.}\vspace{-1em}
	\label{fig:ache}
\end{figure*}

\subsection{Spatial VLM Foundation}
\label{sec:vlm base}
We build a spatial vision–language model following the SR-3D framework \citep{Cheng2025SR3D}, reimplementing its joint text–image architecture. Specifically, the model supports interleaved text, image, and region tokens for spatially grounded reasoning. 
We follow the single-view setup, which applies to single-image or multi-image settings.

For spatial encoding, we adopt SR-3D's positional embedding formulation, which encodes 2D coordinates and the depth map while incorporating camera intrinsics and extrinsics to provide 3D-aware positional bias. 
Then the pixel-wise 3D position map is encoded and added to the corresponding visual embeddings, effectively fusing geometric information into the visual features.

We further introduce the region-prompt interface that allows region tokens (e.g., mask or bounding-box references) to be injected alongside text tokens. 
Through cross-attention, these region tokens interact with vision features, enabling localized reasoning over specific spatial regions. 
Note that the original SR-3D does not include region-level 3D signal prediction or grounding; in the following sections, we extend this capability through additional 2D and 3D grounding supervision.




\subsection{Region-to-3D Detection}
3D spatial localization is a critical capability for spatial reasoning, which enables our detect-then-reason paradigm to perform structured, explicit geometric calculations based on coordinates, yielding more accurate results.
However, directly predicting 3D positions from text is highly challenging.
To address this, we use region tokens as a bridge, leveraging 2D position priors to facilitate 3D localization. 
Instead of localizing directly from language prompts, we input region prompts to the LLM to generate corresponding 3D coordinates, effectively disentangling semantic parsing from 3D perception. 
The region prompt is marked as \texttt{<mask}> in the text to keep consistent with \cite{Cheng2025SR3D}, which is replaced by the visual token of the corresponding image region. 
In Table \ref{tab:ablation dtr}, we show that direct grounding (remove the region prompt) leads to significant performance drop for the DTR inference path.

\subsection{Cold-Start Supervised Fine-tuning}
\label{sec:cold-start}


The whole pipeline is shown in Figure \ref{fig:ache}. We introduce a cold-start phase that equips the base model with essential reasoning capabilities before reinforcement learning. In the following sections, we describe how we construct CoT data for both language-only reasoning (LOR) and detect-then-reason (DTR), and how we blend them with general supervision to initialize the model effectively.

\begin{figure*}[t]
	\centering
	\includegraphics[width=\linewidth]{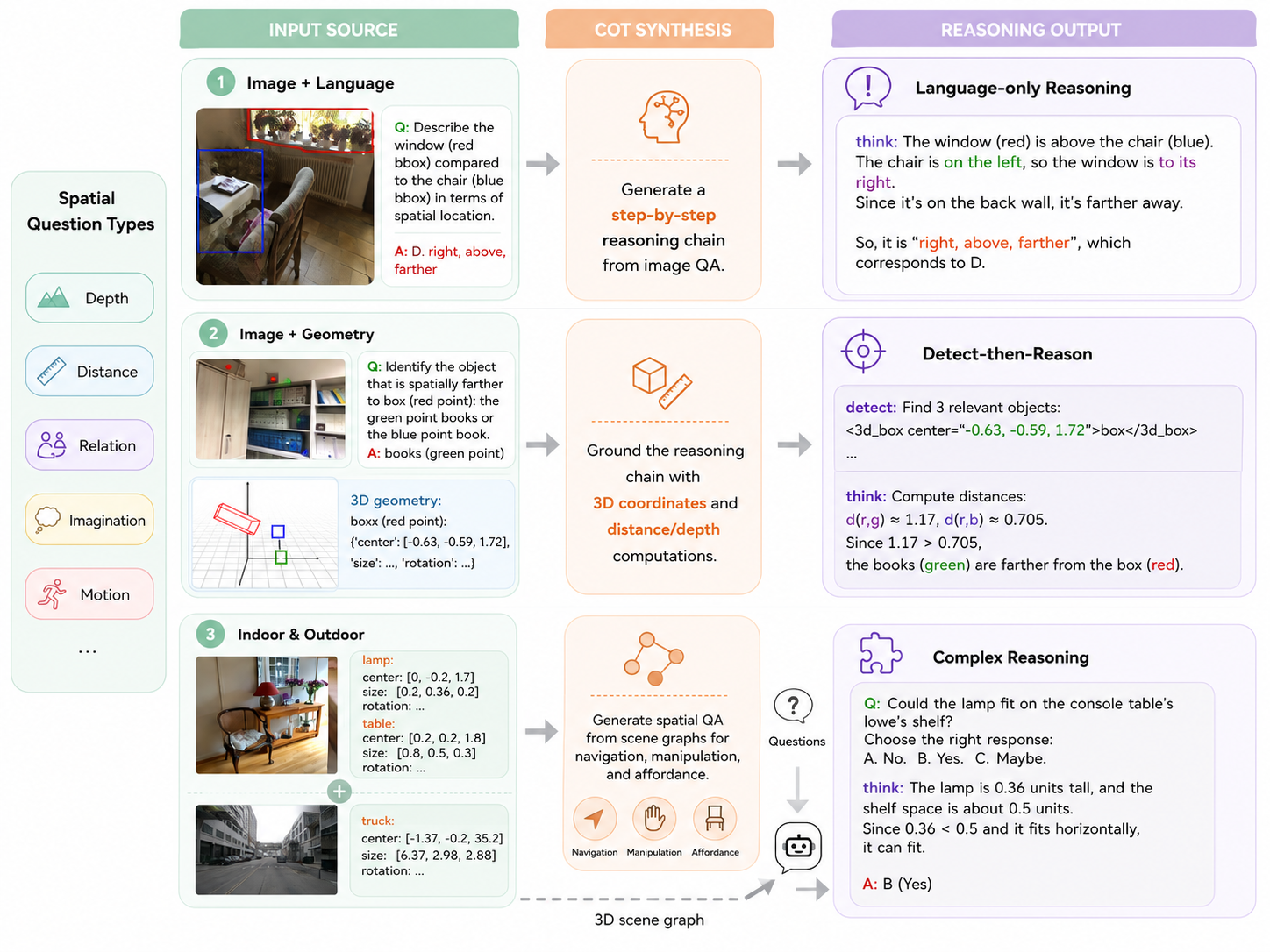}\vspace{-0.2cm}
	\caption{\textbf{CoT Data Construction}: We generate step-by-step CoT of two spatial reasoning paths: Language-Only Reasoning (top), Detect-Then-Reason with geometry-grounded deduction (middle). 
    \textbf{Complex Spatial tasks Construction}: Using multimodal scene-graph datasets that provide both visual and geometric annotations, we prompt LVLM to generate higher-level reasoning task data (bottom).
    }
	\label{fig:data pipe}
\end{figure*}

\noindent\textbf{CoT Data Construction—LOR.}
We first construct language-only CoT data to enhance the model’s ability for textual spatial reasoning. As depicted in the upper part of Figure \ref{fig:data pipe}, given a spatial question and its ground-truth answer, we prompt a LVLM to generate a step-by-step reasoning trace that logically derives the correct answer. The reasoning is required to be concise yet explicit, ensuring that each step connects observed spatial relations (e.g., orientation or distance) to the final decision. Each trace ends with a definitive statement linking reasoning to the answer (e.g., “Therefore, the answer is A (the chair is closer)”). This dataset establishes a linguistic foundation for spatial reasoning without any geometric supervision.

\noindent\textbf{CoT Data Construction—DTR.}
To incorporate explicit geometry, we extend the above process with detected 3D cues. For each spatial question, we either use available 3D annotations or estimate object centers \((x,y,z)\) / 3D bounding boxes. These cues are then integrated into a structured reasoning chain that performs explicit spatial computation (e.g., distances, coordinate comparison, geometric rules) to derive the answer. As shown in the middle part in Figure \ref{fig:data pipe}, for every 2D region annotated in SPAR~\citep{spar}, we retrieve its 3D location by projecting all EmbodiedScan~\citep{DBLP:conf/cvpr/WangMZXLLCZCXLL24/emboddiedscan} 3D object annotations into the image plane using the known camera parameters. 
The matched 3D annotation is then used as the ground-truth 3D coordinate. 
For multi-view questions, we select one frame as main perspective and align each object to the reference coordinate system. 
After prompting the expert model, each output is organized as
``\texttt{<detect> ... </detect><think> ... </think><answer> ... </answer>}''.
Unlike the language-only CoT, this version \emph{enforces} mathematical formulation and deduction. Region tokens from the base spatial VLM’s region branch align text mentions with visual regions (mask/box), supporting accurate spatial grounding during both detection and reasoning.

\noindent\textbf{Complex Spatial Task Generation.}
Beyond template-style spatial questions, we further generate complex spatial tasks involving navigation, object interaction, and layout reasoning, which aims to improve our model's reasoning generalization in diverse spatial scenarios. Using multimodal scene-graph datasets that provide both visual and geometric annotations, we prompt LVLM to generate spatially grounded questions, identify relevant objects, and then produce the corresponding answers and detailed CoT traces. Each example forms a tuple of \texttt{(image, question, coordinates, CoT, answer)}, extending reasoning coverage from simple spatial comparison to higher-level reasoning over structured 3D scenes. 
The data pipeline is shown in the bottom part in Figure \ref{fig:data pipe}.

\noindent\textbf{Quality Control.}
Expert-generated CoT traces may contain hallucinations or invalid reasoning. We apply two-stage filtering: answer matching retains only samples whose conclusion matches the ground truth, and an LLM verifier then checks logical consistency of the rationale and correctness of intermediate spatial computation. We further spot-check sampled batches to verify annotation format and region--object alignment. Data proportions are reported in Appendix~\ref{app:data cot}.



\noindent\textbf{Blended Supervision for Stable Cold-Start.}
Training exclusively on CoT data leads to rapid degeneration of the model’s general multimodal capabilities, including fundamental spatial skills and stable output formatting. 
We observe that without diverse supervision, the model tends to overfit to CoT-style reasoning and shows weaker cross-domain transfer. 
To mitigate this, we introduce a blended fine-tuning stage. 
First, since our spatial VLM backbone lacks grounding capability, we incorporate both 2D and 3D grounding data to build this competence. 
For 2D grounding, we use the RefCOCO \citep{refcoco} dataset, where the model learns to predict 2D bounding boxes given natural language descriptions. 
For the region-to-3D supervision required by DTR, we integrate data from CA-1M, Omni3D, and OmniNOCS \citep{lazarow2024cubify,brazil2023omni3d,krishnan2024omninocs}, enabling the model to output 3D bounding boxes—defined by object centers, dimensions, and orientations.
To further enhance general spatial perception capabilities, we incorporate additional region-prompted spatial fine-tuning data. 
To further preserve linguistic and commonsense priors, we also blend general multimodal SFT data containing standard non-spatial Q\&A pairs. 
Together, these sources form the complete cold-start dataset of roughly one million samples, used to fine-tune the model for two epochs before reinforcement learning.

\subsection{Reinforcement Learning Stage}
\label{sec:grpo}

We train the model with GRPO, a rule-based reinforcement learning method well suited to multiple-choice supervision. Both SPAR and the OpenImages-derived dataset (described later) provide multiple-choice answers, enabling correctness to be used directly as the scalar accuracy reward.

\noindent\textbf{Reward Design.}
Following standard practice in VLM RL, each rollout receives two rewards: a format reward and an accuracy reward. The format reward checks whether the output follows the required ``think–answer'' structure for LOR, and ``detect–think–answer'' structure for DTR. The accuracy reward is computed based on the task type. For multiple-choice questions, a rollout receives a positive score if the selected option matches the ground-truth label and zero otherwise. For filling questions, we use an exponentially smoothed relative error to calculate the reward:
\begin{equation}
    r_{acc}(x) = \exp \left( -2 \cdot \frac{|x - x_{\text{gt}}|}{|x_{\text{gt}}| + \epsilon} \right),
\end{equation}
where $x$ is the predicted value, $x_{\text{gt}}$ is the ground-truth value, and $\epsilon$ is a small constant for numerical stability.

\noindent\textbf{DTR Detection Reward.}
Cold-start learning equips the model with a region-to-3D interface: when region tokens appear in the prompt, the model predicts the corresponding 3D center or bounding box using the learned pattern. RL preserves this output format and evaluates the predicted centers. For each DTR rollout, we extract the predicted 3D centers, compute their distances $d$ to ground-truth annotations, and assign a discretized detection reward:
\begin{equation}\footnotesize
    r_{detect} = \max\left(0, 1 - \left\lfloor \frac{d}{0.2} \right\rfloor \times 0.2\right).
\end{equation}
In multi-view scenarios, we identify which frame the model selects as a reference and calculate the reward with the object's 3D ground-truth center point projected to that specific reference frame.

\noindent\textbf{Online Filtering.}
We apply an online filtering mechanism similar to DAPO: rollout groups whose samples receive identical total reward are removed because they provide no relative advantage. The remaining groups are resampled to maintain the batch size. In this way, training samples have greater advantages, enabling more efficient model training and improving overall performance. 

\noindent\textbf{Training Data.}
The RL stage uses two data sources. (1) \emph{SPAR}: single-view and multi-view questions provide region references and 2D/3D positions for accuracy and detection rewards. (2) \emph{OpenImages-derived data}: to broaden spatial generalization beyond SPAR, we construct an additional dataset based on SRGPT data pipelines. For each OpenImages image, we build a 3D scene graph by combining segmentation masks with monocular depth lifting to approximate per-object point clouds, from which we estimate oriented 3D bounding boxes as detection supervision. 
We then randomly select two or more objects from each scene graph and generate multiple-choice spatial questions describing their geometric relationships with LLM. 
This expanded dataset supplies both global spatial reasoning signals and 3D supervision for DTR rewards, enabling RL to improve linguistic reasoning and geometry-aware detection behavior jointly.

%% file: sec/4_exp.tex
\section{Experiments}
\label{sec:exp}

\begin{table}[t]
\centering
\caption{
\textbf{Results on spatial reasoning benchmarks.} I.e., 
SPAR-Bench, where
\textbf{SR-3D} as its base model,
and \textbf{Ours-LOR} / \textbf{Ours-DTR} denote our final model
evaluated with language-only and detect-then-reason inference, respectively. 
Our method also generalizes and improves the performance of the out-of-domain benchmarks, such as EmbSpatial and SAT.  
SAT contains global scene questions without region grounding, so we use LOR inference only. 
Bold numbers denote the best results in each column. Underlined numbers denote the second-best results.
More detailed results on SPAR-Bench can be found in Appendix \ref{app: more results}.}
\begin{adjustbox}{max width=\textwidth}
\begin{tabular}{lcccccc}
\toprule
& \multicolumn{4}{c}{\textbf{SPAR-Bench}} 
& \multirow{2}{*}{\textbf{EmbSpatial}} 
& \multirow{2}{*}{\textbf{SAT}} \\[-3pt]
& Low & Medium & High & Avg. & & \\ 
\midrule
\rowcolor{gray!10}
\textbf{\textit{General Models}} & & & & & & \\ 
InternVL2.5-8B~\citep{Chen2023InternVL} & 21.7 & 31.1 & 36.3 & 29.7 & 59.8 &  57.3 \\
LLaVA-OneVision-1.5-8B~\citep{DBLP:journals/corr/abs-2408-03326/llava_ov} & 31.9 & 26.0 & 42.4 & 35.5 & 67.2  &   64.0 \\
Qwen2.5-VL-7B~\citep{DBLP:journals/corr/abs-2502-13923/qwen25vl} & 17.5 &  29.5  & 41.8 & 30.2 & 70.4 & 62.0 \\
Qwen3-VL-8B~\citep{DBLP:journals/corr/abs-2511-21631/qwen3vl} & 35.7 &  30.0 & 46.7 & 39.6 & 79.0  & \textbf{69.3} \\
SpatialRGPT~\citep{Cheng2024SpatialRGPT} & 26.0  & 22.0 & 32.1 & 28.0 & 60.9  & 44.0 \\
\rowcolor{blue!10}
NVILA-8B-Lite (Base)~\citep{liu2025nvila} & 24.9 & 28.6 & 40.3 & 32.3 & 67.3  &   63.3 \\
\midrule
\rowcolor{gray!10}
\textbf{\textit{Spatial Reasoning Models}} & & & &  &  & \\
Cosmos-Reason1-7B~\citep{lin2025cosmosreason1} &  23.4 & 19.4 & 24.2 & 22.8 & 65.2 &  60.7 \\
ViGoRL~\citep{sarch2025vigorl} & 25.0 & 18.1 & 18.6 &  21.1 & 71.8 &   61.3 \\
SpatialLadder~\citep{li2025spatialladder} &  25.8 & 33.2 & 42.6  &  34.4 & 59.8 &   16.7 \\
SpaceR~\citep{ouyang2025spacer} & 32.3 & 33.0 & 47.6 & 39.2 &  69.4 &   64.7 \\
VST~\citep{DBLP:journals/corr/abs-2511-05491/vst} & 53.3 & 25.4 & 53.7 & 48.9 & 70.4  & \textbf{69.3} \\
\midrule
\rowcolor{gray!10}
\textbf{\textit{Ours}} &  & & &  &  & \\
\rowcolor{blue!10}
SR-3D (Base)~\citep{Cheng2025SR3D} & 24.1 & 37.6 & 40.1 & 33.4 & 72.5 &  63.0 \\
\textbf{Ours-LOR} & \underline{58.5} & \textbf{47.7} & \underline{67.3} & \underline{60.5} & \underline{79.2}  &   \underline{68.7} \\
\textbf{Ours-DTR} & \textbf{61.1} & \underline{46.9} & \textbf{68.3}  & \textbf{61.9} & \textbf{81.3} &  - \\
\bottomrule
\end{tabular}
\end{adjustbox}
\label{tab:main-res1}
\end{table}

\begin{table}[t]
\centering
\caption{
\textbf{Comparisons of LOR/DTR inference on representative SPAR-Bench subtasks for single-view and multi-view scenarios.} We report accuracies on \texttt{depth\_prediction\_oo} (Depth), \texttt{distance\_infer\_center\_oo} (Distance), \texttt{obj\_spatial\_relation\_oo} (Relation), and \texttt{spatial\_imagination\_oo} (Imagination).
}
\begin{adjustbox}{max width=\textwidth}
\begin{tabular}{ccccccccc}
\toprule
         & \multicolumn{4}{c}{\textbf{Single-View}}           & \multicolumn{4}{c}{\textbf{Multi-view}}            \\
         & Depth & Distance & Relation & Imagination & Depth & Distance & Relation & Imagination \\
         \midrule
Ours-LOR & 30.5  & 75.0     & 71.4     & 49.3        & 25.9  & \textbf{67.0}     & \textbf{78.9}     & 69.5        \\
Ours-DTR & \textbf{35.4}  & \textbf{80.0}     & \textbf{73.9}     & \textbf{52.0}        & \textbf{30.4}  & \textbf{67.0}     & 77.8     & \textbf{69.7}     \\
\bottomrule
\end{tabular}
\end{adjustbox}
\label{tab:spar-result}
\end{table}

\begin{table}[t]
\centering
\caption{Results on \textbf{OOD} benchmarks: BLINK (Spatial), RealWolrdQA, and CVBench.}
\begin{adjustbox}{max width=\textwidth}
\begin{tabular}{l|cccc}
\toprule
             & BLINK(s)  & RWQA  & CVBench\\ \midrule
SR-3D (base)       & 83.9     & 68.1 & 88.9 \\
Ours-LOR (Full) & 80.4       &   59.5  &  88.5 \\
Ours-direct (Full)   & \textbf{87.4}       &  64.6  & 88.1 \\
\bottomrule
\end{tabular}
\end{adjustbox}
\label{tab:ood res}
\end{table}

\subsection{Experimental Setup}

\noindent\textbf{Training Data--Cold-Start.}
We train a single unified model from SR-3D that supports both reasoning modes (LOR and DTR). 
The Cold-Start stage is built from four components, totaling approximately 1M samples. Each component is summarized below:
\begin{itemize}
    \item \textbf{CoT--LOR.} 
    We construct 30k CoT samples following the LOR reasoning pathway. SPAR~\citep{spar} is used as the primary source, providing 10k concise reasoning traces on single-view and multi-view data. The remaining 20k are drawn for complex spatial task reasoning construction, including indoor CA-1M~\citep{lazarow2024cubify} and outdoor NuScenes~\citep{nuscenes2020} scenes.
    \item \textbf{CoT--DTR.}
    We curate 10k CoT samples requiring explicit region-based detection followed by quantitative analysis and reasoning. These samples are also derived from SPAR, aligning with the DTR pathway used later in RL optimization.
    \item \textbf{Grounding Data.}
    To support spatial localization, we include 3D grounding from Omni3D~\citep{brazil2023omni3d}, OmniNOCS~\citep{DBLP:conf/eccv/KrishnanKMHB24/omninocs} and 2D grounding from RefCOCO~\citep{refcoco}. These examples provide region-to-3D point supervision crucial for DTR.
    \item \textbf{Region-Prompted VQA.}
    We incorporate region-aware QA pairs from SRGPT~\citep{Cheng2024SpatialRGPT}, which help the model build localized spatial understanding tied to visual regions.
    \item \textbf{General-Purpose VQA.}
    To maintain broad multimodal capability and prevent overfitting to spatial reasoning, we include non-spatial QA data from LLaVA-1.5~\citep{Liu2024LLaVA15}.
\end{itemize}

\noindent\textbf{TrainingData--RL.}
The RL stage uses roughly 200k spatial questions spanning single-view and multi-view settings, in both multiple-choice and filling formats: $\sim$100k global questions (LOR) and $\sim$100k region-grounded questions with 2D/3D object coordinates (DTR), sourced from SPAR and an OpenImages-derived dataset (see Section~\ref{sec:grpo}).

\noindent\textbf{Evaluation Benchmarks.} Our model supports both LOR and DTR inference paradigms \textit{on the same checkpoint}. 
(i) \textbf{Region-Grounded} benchmarks consist of questions tied to specific image regions, allowing us to evaluate both LOR and DTR inference. We use \emph{SPAR-Bench}~\citep{spar}, which contains multiple-choice and filling questions on single-view and multi-view scenarios, covering 20 subtasks including distance, spatial relationships, and viewpoint changes. We also include \emph{EmbSpatial-Bench}~\citep{du-etal-2024-embspatial}, which specifically targets positional relationships in embodied environments.
(ii) \textbf{Global-Only} benchmarks contain only image-level questions without region grounding, such as the dynamic spatial benchmark \emph{SAT}~\citep{ray2025satdynamicspatialaptitude}; here we use LOR inference only. 
(iii) \textbf{OOD} benchmarks evaluate robustness under distribution shift. We include \emph{BLINK}~\citep{fu2024blink} for perception-heavy tasks, \emph{RealWorldQA}~\citep{realworldqa_xai2024} for real-world driving and everyday scenes, and \emph{CVBench}~\citep{tong2024cambrian} which focuses on spatial counting and relationships.


\subsection{Main Results}

\noindent\textbf{Performance on Standard Benchmarks.}
We first evaluate our model on standard spatial benchmarks including SPAR-Bench, EmbSpatial, and SAT, as shown in Table~\ref{tab:main-res1} and Table~\ref{tab:spar-result}. 
On \textbf{SPAR-Bench}, which features diverse question types (single/multi-image, multiple-choice/filling), our model achieves significant improvements over the base model.
Specifically, Ours-DTR attains an average accuracy of 61.9, surpassing the base SR-3D (33.4) by a large margin (+28.5). 
Similarly, on \textbf{EmbSpatial}, which focuses on embodied positional relationships, Ours-DTR reaches 81.3 compared to 72.5 for the base. 
On the global benchmark \textbf{SAT}, where no region prompts are available, our model using LOR (68.7) maintains better than the base (63.0), showing that RL tuning preserves global spatial reasoning.
Table~\ref{tab:spar-result} further details the performance on representative subtasks of SPAR-Bench.
Across both single-view and multi-view settings, the geometry-aware DTR mode consistently outperforms LOR on quantitative tasks, such as \emph{Depth Prediction} (+4.9 on Single-View) and \emph{Distance Inference} (+5.0 on Single-View). 
This confirms that accessing explicit 3D coordinates enables more precise calculation compared to purely linguistic reasoning. 
For \emph{Spatial Relation} and \emph{Imagination} tasks, DTR remains highly competitive, often exceeding LOR, demonstrating that explicit detection benefits general spatial reasoning as well.

\noindent\textbf{Generalization on OOD Benchmarks.} 
We further assess the model’s robustness on out-of-distribution (OOD) benchmarks—\emph{BLINK(s)}, \emph{RealWorldQA}, and \emph{CVBench}—which involve unseen imagery and question styles (Table~\ref{tab:ood res}). Although trained on region-grounded spatial QA, the model maintains strong generalization under these shifts. 
While applying CoT reasoning to these distributions leads to lower performance, we find that the model maintains high accuracy when using ``direct inference'' (i.e., answering directly without CoT). 
For example, direct inference achieves 87.4\% on BLINK(s) and 88.1\% on CVBench, significantly surpassing the CoT mode. 
This result confirms that our training framework enhances complex spatial reasoning without catastrophic forgetting, preserving the model's original strong capabilities for general spatial perception and QA tasks.

\begin{figure*}[t]
	\centering
	\includegraphics[width=\linewidth]{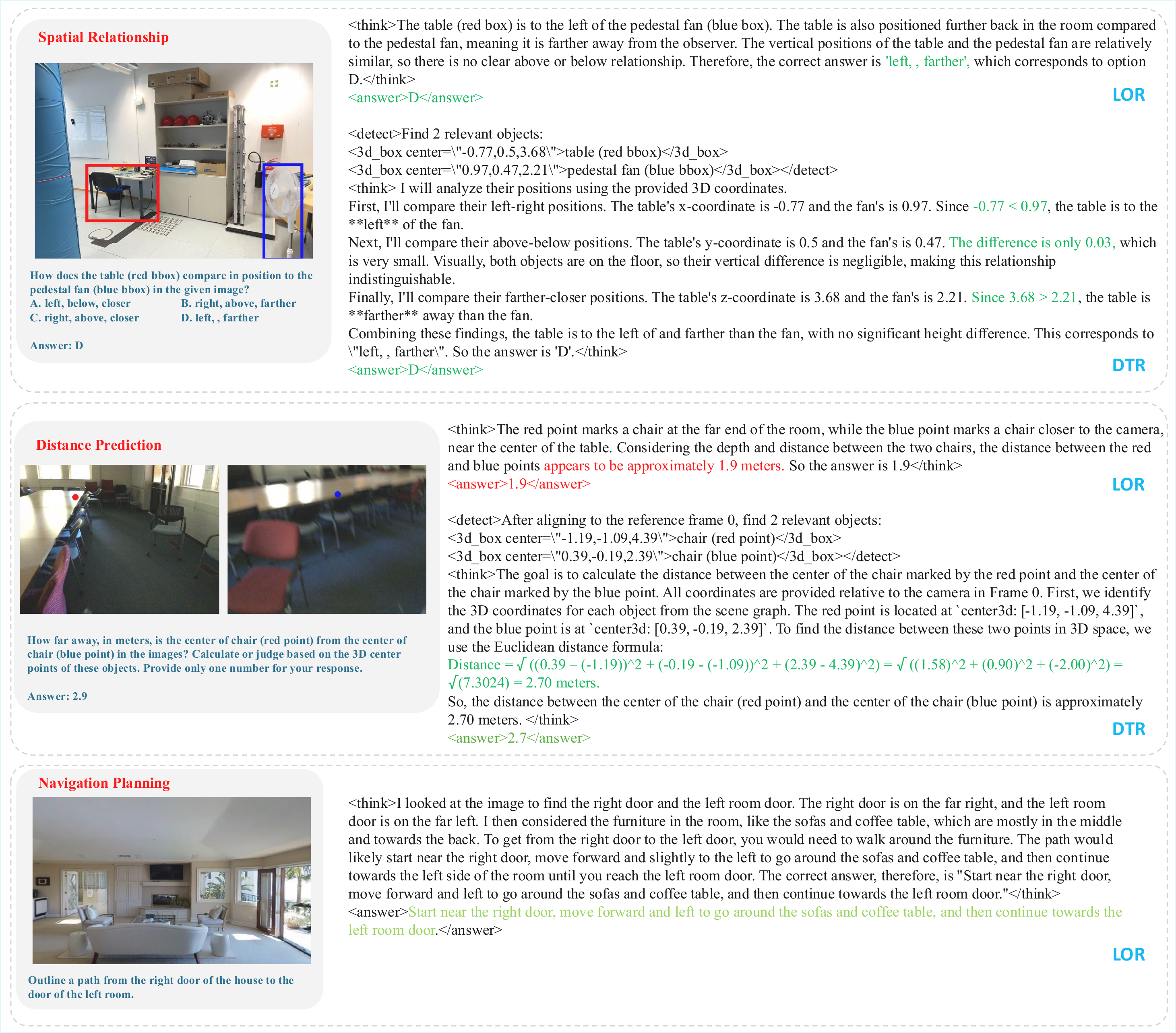}
	\caption{Visualization examples of our model. On the fundamental spatial question (spatial relationship and distance), we compare the reasoning paths of LOR and DTR. On the complex spatial task (navigtion), our model also demonstrates reasoning generalization capability.}
	\label{fig:examples}
\end{figure*}


\begin{table}[t]
\centering
\caption{\textbf{Ablation on LOR and DTR training.} We compare models trained with LOR-only, DTR-only, and mixed data. Columns indicate the inference mode used during evaluation (LOR/DTR).}
\begin{adjustbox}{max width=\textwidth}
\begin{tabular}{c|cccc}
\toprule
                           & \multicolumn{2}{c}{\textbf{SPAR-Bench}} & \multicolumn{2}{c}{\textbf{EmbSpatial}} \\ 
                           & -LOR           & -DTR             & -LOR          & -DTR         \\   \midrule
Ours (LOR)                  &       58.0      & -               &     75.9       & -           \\
Ours (DTR)             & -             &    57.2    & -           &     71.4       \\
Ours (Full)   &       \textbf{58.7}         &        \textbf{60.8}        &        \textbf{77.6}        &    \textbf{78.8}        \\
\bottomrule
\end{tabular}
\end{adjustbox}
\label{tab:ablation reason}
\end{table}

\begin{table}[t]
\centering
\caption{\textbf{Ablation on DTR designs.} We validate the impact of RL detection reward and region-to-3D mechanisms. All metrics are evaluated under DTR inference. \textbf{SPAR-Det} denotes 3D localization error on a validation set of SPAR.}
\begin{adjustbox}{max width=\textwidth}
\begin{tabular}{c|ccc}
\toprule
                    & \textbf{SPAR-Bench} & \textbf{EmbSpatial}  &    \textbf{SPAR-Det}    \\   \midrule
w/o detect reward        &       59.9       &       76.0     &  0.78       \\
w/o region-to-3D          &    59.3    &    74.8      &      0.67      \\
Ours    &      \textbf{60.6}         &      \textbf{78.5}     &  \textbf{0.45} \\
\bottomrule
\end{tabular}
\end{adjustbox}
\label{tab:ablation dtr}
\end{table}

\begin{table}[t]
\centering
\caption{Effect of \textbf{Cold-Start/RL stages.} The ablation is conducted with LOR-only training on selection questions. 
``Cold-Start only'' removes RL, ``RL only'' skips cold-start, and 
``Full pipeline'' uses both stages. ``SI'': single-iview, ``MI'': multi-view.}
\begin{adjustbox}{max width=\textwidth}
\begin{tabular}{lcccc}
\toprule
                     & SPAR(SI) & SPAR(MI) & SAT    & EmbSpatial \\ \midrule
SR-3D (Base)         & 38.97    & 41.49    & 63.00  & 72.50      \\
Cold-Start only      & 56.53    & 45.89    & 62.67  & 65.66      \\
RL only   & 62.34    & 64.22    & \textbf{65.33}  & 76.75      \\
Full pipeline  & \textbf{72.21}    & \textbf{69.39}    & 64.67  & \textbf{76.90}      \\
\bottomrule
\end{tabular}
\end{adjustbox}
\label{tab:stage}
\end{table}

\begin{table}[t]
\centering
\caption{\textbf{Ablation of Cold-Start training data} in LOR training setup.}
\begin{adjustbox}{max width=\textwidth}
\begin{tabular}{cccc|cc}
\toprule
  $\text{CoT}_{\text{spar}}$   &   $\text{CoT}_{\text{CA}}$   &  MM-data     &     Region-data      & SPAR(SI) & EmbSpatial \\   \midrule
 $\checkmark$  &  $\times$    &    $\times$   &     $\times$  &          69.30       &      64.34          \\
$\checkmark$ &  $\checkmark$    &   $\times$   &    $\times$   &        67.13         &          68.81      \\
$\checkmark$ &  $\checkmark$  &  $\checkmark$  &    $\times$     & 69.16           &       76.55    \\
$\checkmark$  &  $\checkmark$  &  $\checkmark$  &  $\checkmark$  & \textbf{72.21}           &  \textbf{76.90}    \\   
\bottomrule
\end{tabular}
\end{adjustbox}
\label{tab:cold start}
\end{table}

\subsection{Qualitative Results}

Figure \ref{fig:examples} shows visualization examples of our model's reasoning. 
In the first case of spatial relation task, our model can reason effectively through either linguistic deduction or explicit coordinate comparison. 
The second case involves a challenging multi-view distance estimation problem. 
LOR relies on visual estimation and fails to gauge the metric distance accurately across viewpoints. 
In contrast, DTR aligns the chair from the second view to the first view's reference frame, localizes both objects within the same coordinate system, and computes the distance explicitly, yielding a precise answer. 
The final example involves multi-step planning for a complex navigation task. Our model first localizes the start and end points, then identifies obstacles along the path, and finally generates the correct route plan. This capability is primarily attributed to the CoT data for complex problems constructed from CA-1M and NuScenes during the cold-start phase, enabling the model to handle more sophisticated spatial tasks. 
More additional examples are provided in the Appendix \ref{app:visualization}.

\subsection{Ablation Analysis}

\noindent\textbf{Impact of Joint LOR \& DTR Training.} 
Table~\ref{tab:ablation reason} investigates the benefits of unifying LOR and DTR training. 
We compare our full model against variants trained exclusively on LOR or DTR data. 
The results show that joint training not only supports both inference modes within a single model but also fosters mutual reinforcement: 
the model trained on both (Ours-Full) consistently outperforms the single-mode baselines across all metrics. 
For instance, LOR inference improves from 58.0 to 58.7 on SPAR-Bench when DTR data is added, suggesting that learning explicit geometry enhances the model's underlying spatial representation. 
Conversely, DTR inference benefits from LOR training (improving from 57.2 to 60.8), likely because pure DTR training can lead to an over-reliance on quantitative calculation at the expense of qualitative spatial perception and reasoning. 
This confirms our hypothesis that LOR and DTR are complementary paradigms, and a unified model can effectively leverage both.

\noindent\textbf{Ablation on DTR Designs.} 
Table~\ref{tab:ablation dtr} validates the critical components of our DTR mechanism: the RL detection reward and the region-to-3D interface. 
Here, we evaluate all models using DTR inference and report 3D localization error on a validation set (\texttt{SPAR-Det}) consisting 400 instances. 
Removing the discrete detection reward leads to a significant drop in localization accuracy (error increases from 0.45 to 0.78) and a corresponding decline in reasoning performance, demonstrating that precise 3D grounding is essential for correct geometric deduction. 
Similarly, ablating the region-to-3D interface (i.e., predicting coordinates directly from text without visual region tokens) degrades performance, highlighting the significance of 2D position priors. 
These results confirm that both explicit supervision for intermediate detection steps and the region-to-3D mechanism are vital for the success of the detect-then-reason paradigm.

\noindent\textbf{Impact of Cold-Start SFT and RL Phases.} 
Table \ref{tab:stage} disentangles the effects of the SFT and RL stages in LOR training. After cold-start SFT, SPAR-Bench performance increases due to the CoT data from SPAR, while metrics on unseen tasks like SAT and EmbSpatial decline. Adding the RL stage improves both reasoning and generalization, enabling the model to solve EmbSpatial and achieving the strongest results on SPAR and EmbSpatial. We also test an ``RL only’’ variant that applies RL directly to the spatial base model. Although it improves metrics and even achieves the best SAT performance, its generated CoTs are often illogical or inconsistent with the answers, indicating that RL alone cannot bootstrap coherent reasoning (see Appendix \ref{app:visualization} for examples).

\noindent\textbf{Analysis of Cold-Start Data Components.}
In Table \ref{tab:cold start}, each data group contributes positively to the final performance. Cold-starting with only SPAR CoT data yields strong SPAR-Bench results but poor generalization, including a low 64.34 on EmbSpatial and occasional chaotic reasoning on OOD tasks. Adding CoT data from other sources (e.g., CA-1M) and general-purpose multimodal data progressively improves generalization, leading to large gains on EmbSpatial. Incorporating region-related fine-tuning data further boosts performance and yields the best results across all benchmarks. Thus, diverse data sources during cold start are essential for strong generalization and spatial perception.

%% file: sec/5_conl.tex
\section{Conclusion}\vspace{-2mm}
SR-REAL offers a novel approach to spatial reasoning in Vision–Language Models, demonstrating the effectiveness of reinforcement learning in enhancing both linguistic and geometric reasoning. Our framework integrates two complementary reasoning paths, LOR and DTR, leveraging region-to-3D grounding and discrete detection rewards to improve spatial accuracy. SR-REAL excels across a range of spatial benchmarks and achieves strong cross-domain generalization. 

%% file: sec/X_suppl.tex
\clearpage
\appendix
\setcounter{page}{1}



\section{Implementation Details}
\label{app:impl}

\noindent\textbf{Stage 1: Cold-Start SFT.}
We fine-tune the SR-3D base model on the blended cold-start dataset ($\sim$1M samples) for 2 epochs. Training uses a learning rate of $5\times10^{-6}$ with cosine decay scheduling and a batch size of 128.

\noindent\textbf{Stage 2: Reinforcement Learning.}
We apply GRPO-based RL for 200 steps on the $\sim$200k spatial RL dataset. The rollout batch size is 512 and the learning rate is $1\times10^{-6}$ with cosine decay. Both stages are trained on 32 NVIDIA A100 GPUs.

\section{Preliminary knowledge of GRPO}
\label{app: grpo}

In this work, we employ Group Relative Policy Optimization (GRPO) as our core reinforcement learning algorithm. GRPO is a policy optimization method that eliminates the need for a value function critic—common in algorithms like PPO—thereby reducing memory usage and computational overhead during training.

The fundamental principle of GRPO involves estimating the advantage of a sampled response by comparing its reward against a group of other responses generated from the same input query. Formally, for a given question-answer pair query $q$, the old behavior policy $\pi_{\theta_{\text{old}}}$ samples a group of $G$ outputs $\{o_i\}_{i=1}^G$. The optimization process consists of the following key components:

\noindent\textbf{Group-Relative Advantage Estimation.} Instead of relying on a learned value function to predict the baseline, GRPO calculates the advantage $\hat{A}_{i,t}$ for the $i$-th response by normalizing its reward $r_i$ with respect to the group's statistics. The advantage for each token $t$ in the response $o_i$ is defined as:

\begin{equation}
\hat{A}_{i,t} = \frac{r_i - \text{mean}(\{r_i\}_{i=1}^G)}{\text{std}(\{r_i\}_{i=1}^G)},
\end{equation}

\noindent where $r_i$ is the reward obtained for the $i$-th output. This normalization effectively serves as a baseline, encouraging the model to reinforce outputs that perform better than the group average.

\noindent\textbf{Objective Function.} GRPO maximizes a surrogate objective that incorporates importance sampling ratios and the PPO-style clipping mechanism to ensure stable updates. Additionally, a Kullback-Leibler (KL) divergence penalty is added directly to the objective to prevent the trained policy $\pi_\theta$ from deviating excessively from the reference policy $\pi_{\text{ref}}$. The full objective function is given by:

\begin{equation}
\begin{split}
\mathcal{J}_{\text{GRPO}}(\theta) = & \mathbb{E}_{q \sim \mathcal{D}, \{o_i\}_{i=1}^G \sim \pi_{\theta_{\text{old}}}(\cdot|q)} \Bigg[ \frac{1}{G} \sum_{i=1}^G \frac{1}{|o_i|} \sum_{t=1}^{|o_i|} \\
& \bigg( \min \left( \rho_{i,t}(\theta) \hat{A}_{i,t}, \text{clip}(\rho_{i,t}(\theta), 1-\epsilon, 1+\epsilon) \hat{A}_{i,t} \right) \\
& - \beta D_{\text{KL}}(\pi_\theta || \pi_{\text{ref}}) \bigg) \Bigg],
\end{split}
\end{equation}

\noindent where $\rho_{i,t}(\theta)$ is the probability ratio between the current and old policies:
\begin{equation}
\rho_{i,t}(\theta) = \frac{\pi_\theta(o_{i,t} \mid q, o_{i,<t})}{\pi_{\theta_{\text{old}}}(o_{i,t} \mid q, o_{i,<t})}.
\end{equation}

Here, $\epsilon$ is the clipping parameter, $\beta$ is the coefficient for the KL penalty, and $D_{\text{KL}}$ measures the divergence between the current policy and the reference policy at the token level. This formulation allows GRPO to efficiently optimize the policy using group-based relative feedback without maintaining a separate critic model.

\section{CoT Data Construction}
\label{app:data cot}

\begin{figure*}[t]
	\centering
	\includegraphics[width=\linewidth]{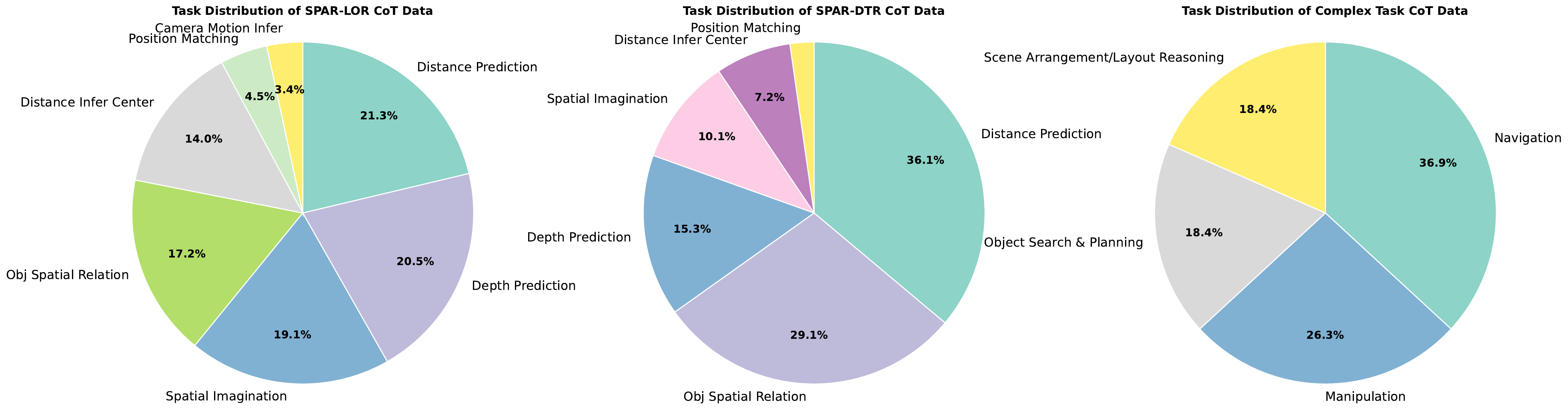}
	\caption{\textbf{Task distribution of CoT instruction data.} The LOR and DTR subsets (10k each) are sourced from SPAR and cover fundamental spatial tasks including depth, distance, spatial relations, spatial imagination, and position matching. The complex task subset (20k) is drawn from CA-1M and NuScenes and covers higher-level scenarios such as navigation, object manipulation, and spatial planning.
    }
	\label{fig:cot data distribution}
\end{figure*}

\noindent\textbf{CoT Data Composition.}
Figure~\ref{fig:cot data distribution} illustrates the task distribution of our CoT instruction data. We construct approximately 10k LOR samples and 10k DTR samples sourced from the SPAR dataset, covering fundamental spatial perception tasks including depth estimation, distance prediction, object spatial relations, spatial imagination, and position matching. In addition, we construct 20k complex spatial task CoT samples drawn from the CA-1M and NuScenes datasets, targeting higher-level reasoning scenarios such as navigation, object manipulation, and spatial planning. Together, these three components provide balanced coverage across both basic geometric reasoning and complex scene-level inference.

\noindent\textbf{Prompts.} 
Our model supports two distinct inference pathways, where the specific reasoning mode is determined by the input prompt. The system prompts corresponding to Language-Only Reasoning (LOR) and Detect-Then-Reason (DTR) are presented in Table \ref{tab:sys prompt sr-real}. In contrast to LOR, the DTR paradigm explicitly instructs the model to output 3D detection information enclosed within a $<$detect$>$ block prior to thinking. Furthermore, Tables \ref{tab:gemini prompt LOR}, \ref{tab:gemini prompt DTR}, and \ref{tab:sys prompt questions} illustrate the system prompts used to query the LVLM during the construction of LOR CoT data, DTR CoT data, and complex spatial reasoning scenarios, respectively.

\begin{table}[t]
\centering
\caption{
\textbf{Detailed results on SPAR-Bench.}
We report scores of 20 dimensions, together with the overall average. 
}
\begin{adjustbox}{max width=\textwidth}
\scriptsize
\setlength{\tabcolsep}{2.6pt}
\begin{tabular}{lccccccccccccccccccccc}
\toprule
\textbf{Model} & \rotatebox{90}{\texttt{camera\_motion\_infer}} 
& \rotatebox{90}{\texttt{depth\_prediction\_oc}} 
& \rotatebox{90}{\texttt{depth\_prediction\_oc\_mv}} 
& \rotatebox{90}{\texttt{depth\_prediction\_oo}} 
& \rotatebox{90}{\texttt{depth\_prediction\_oo\_mv}} 
& \rotatebox{90}{\texttt{distance\_infer\_center\_oo}} 
& \rotatebox{90}{\texttt{distance\_infer\_center\_oo\_mv}} 
& \rotatebox{90}{\texttt{distance\_prediction\_oc}} 
& \rotatebox{90}{\texttt{distance\_prediction\_oc\_mv}} 
& \rotatebox{90}{\texttt{distance\_prediction\_oo}} 
& \rotatebox{90}{\texttt{distance\_prediction\_oo\_mv}} 
& \rotatebox{90}{\texttt{obj\_spatial\_relation\_oc\_mv}} 
& \rotatebox{90}{\texttt{obj\_spatial\_relation\_oo}} 
& \rotatebox{90}{\texttt{obj\_spatial\_relation\_oo\_mv}} 
& \rotatebox{90}{\texttt{position\_matching}} 
& \rotatebox{90}{\texttt{spatial\_imagination\_oc}} 
& \rotatebox{90}{\texttt{spatial\_imagination\_oc\_mv}} 
& \rotatebox{90}{\texttt{spatial\_imagination\_oo}} 
& \rotatebox{90}{\texttt{spatial\_imagination\_oo\_mv}} 
& \rotatebox{90}{\texttt{view\_change\_infer}} 
& \rotatebox{90}{\textbf{Avg.}} \\
\midrule
Gemini-2.5-Pro & 28.5 & 56.1 & 42.7 & 18.3 & 16.2 & 73.2 & 64.3 & 49.5 & 53.3 & 44.8 & 38.0 & 52.5 & 71.2 & 74.8 & 35.6 & 39.5 & 43.0 & 30.8 & 65.0 & 15.6 & 45.4 \\
InternVL2.5-8B~\cite{Chen2023InternVL} & 26.2 & 26.2 & 27.2 & 12.1 & 15.1 & 57.6 & 52.7 & 24.8 & 25.2 & 22.3 & 21.4 & 39.5 & 35.7 & 30.2 & 57.0 & 25.0 & 29.7 & 25.5 & 30.8 & 10.7 & 29.7 \\
LLaVA-OneVision-1.5-8B~\cite{DBLP:journals/corr/abs-2408-03326/llava_ov} & 29.8 & 41.8 & 35.1 & 19.1 & 16.2 & 62.4 & 58.9 & 43.6 & 42.1 & 30.7 & 25.4 & 43.8 & 45.1 & 48.2 & 43.5 & 30.1 & 34.3 & 22.8 & 35.0 & 5.1 & 35.5 \\
Qwen3-VL-8B~\cite{DBLP:journals/corr/abs-2511-21631/qwen3vl} & 25.2 & 51.4 & 54.7 & 17.4 & 17.5 & 70.3 & 67.9 & 22.8 & 37.2 & 50.5 & 38.9 & 50.8 & 63.7 & 40.4 & 48.6 & 28.2 & 32.6 & 34.4 & 31.7 & 16.4 & 39.6 \\
NVILA-8B-Lite~\cite{liu2025nvila} & 25.8 & 27.4 & 33.9 & 19.8 & 17.2 & 60.0 & 58.3 & 27.6 & 26.9 & 24.0 & 23.5 & 29.2 & 39.8 & 49.0 & 53.4 & 25.5 & 33.7 & 23.5 & 44.5 & 7.1 & 32.3 \\
SpatialRGPT-VILA1.5-8B~\cite{Cheng2024SpatialRGPT} & 26.2 & 29.5 & 29.3 & 17.3 & 17.1 & 51.8 & 53.3 & 23.4 & 29.3 & 32.1 & 32.0 & 39.8 & 18.4 & 27.4 & 26.2 & 27.7 & 27.9 & 17.5 & 24.9 & 13.6 & 28.0 \\
RynnBrain~\cite{DBLP:journals/corr/abs-2602-14979/rynnbrain} & 30.5 & 70.3 & 55.5 & 11.3 & 14.2 & 82.4 & 81.8 & 68.2 & 62.7 & 51.1 & 36.5 & 44.2 & 63.7 & 42.9 & 53.9 & 34.1 & 36.0 & 29.1 & 33.9 & 10.6 & 45.4 \\
ViGoRL~\cite{sarch2025vigorl} & 1.2 & 40.4 & 30.0 & 17.7 & 17.7 & 60.0 & 31.5 & 34.8 & 32.1 & 13.3 & 12.3 & 5.5 & 29.9 & 5.3 & 34.1 & 16.9 & 0.0 & 20.2 & 2.2 & 19.1 & 21.1 \\
SpaceR~\cite{ouyang2025spacer} & 40.0 & 39.9 & 40.7 & 17.9 & 15.4 & 64.7 & 55.4 & 45.4 & 41.7 & 28.7 & 28.8 & 58.8 & 54.1 & 50.4 & 39.7 & 32.8 & 41.6 & 27.2 & 40.9 & 19.5 & 39.2 \\
VST~\cite{DBLP:journals/corr/abs-2511-05491/vst} & 34.0 & 72.5 & 54.8 & 33.4 & 22.9 & \textbf{88.2} & \textbf{82.4} & 73.0 & 66.3 & 63.3 & 37.6 & 49.0 & 66.8 & 47.4 & 31.6 & 39.8 & 43.3 & 30.1 & 36.7 & 10.7 & 48.9 \\
SR-3D (Base)~\cite{Cheng2025SR3D} & 30.8 & 25.6 & 36.5 & 15.5 & 12.3 & 56.2 & 56.8 & 19.8 & 26.8 & 28.2 & 32.0 & 46.8 & 45.3 & 42.1 & 54.5 & 28.2 & 27.0 & 25.2 & 32.2 & 27.9 & 33.4 \\
\textbf{Ours-LOR} & \textbf{43.8} & 80.4 & \textbf{69.2} & 30.5 & 25.9 & 75.0 & 67.0 & 78.1 & 70.4 & 63.6 & 49.7 & \textbf{80.5} & 71.4 & \textbf{78.9} & \textbf{70.5} & \textbf{55.6} & 54.4 & 49.3 & 69.5 & \textbf{29.2} & 60.5 \\
\textbf{Ours-DTR} & \textbf{43.8} & \textbf{81.1} & 69.1 & \textbf{35.4} & \textbf{30.4} & 80.0 & 67.0 & \textbf{79.0} & \textbf{72.5} & \textbf{65.9} & \textbf{55.2} & \textbf{80.5} & \textbf{73.9} & 77.8 & 68.2 & 54.8 & \textbf{55.5} & \textbf{52.0} & \textbf{69.7} & 29.1 & \textbf{61.9} \\

\bottomrule
\end{tabular}
\end{adjustbox}
\label{tab:spar-dimension-results}
\end{table}

\section{More Results}
\label{app: more results}

\noindent\textbf{Detailed results on SPAR-Bench.}
Table~\ref{tab:spar-dimension-results} reports per-dimension scores across all 20 subtasks on SPAR-Bench, complementing the aggregated results presented in the main paper. Ours-DTR achieves the best overall average (61.9) and leads on most depth- and distance-related dimensions, consistent with its explicit 3D grounding capability. Ours-LOR performs competitively across relation and spatial imagination tasks. Both modes substantially outperform the SR-3D base model across nearly all dimensions.

\noindent\textbf{Training mechanisms during RL.} 
In Table \ref{tab: filter kl}, we analyze the impact of standard GRPO enhancement mechanisms—specifically the online filter (as detailed in the main text) and KL coefficient decay—on reinforcement learning for spatial tasks. 
KL coefficient decay targets the KL divergence term in the GRPO objective, which acts as a constraint to keep the updated policy close to the reference model. During training, we anneal this coefficient following a cosine decay schedule. 
Comparing Rows 2 and 3, we observe that incorporating the online filter not only boosts the model's final performance but also significantly enhances training efficiency: the model converges to the reported results in just 120 steps, whereas standard RL requires 300 steps. 
Furthermore, a comparison of the last two rows reveals that while KL coefficient decay leads to a performance drop on SPAR-BENCH, it yields improvements on benchmarks that diverge from the training distribution, such as SAT and EmbSpatial, thereby enhancing generalization.

\noindent\textbf{Grounding data.}
Table \ref{tab: ground data} illustrates the critical role of grounding data during the cold-start phase. As shown in the second row, the absence of auxiliary grounding supervision leads to inaccurate localization, resulting in performance degradation across all benchmarks under DTR inference. This decline is particularly pronounced on EmbSpatial and CVBench. However, the impact on SPAR-Bench is relatively minor; this is because the cold-start dataset already includes DTR CoT samples derived specifically from SPAR, which provide implicit localization guidance even without separate grounding data.

\section{More Visualization}
\label{app:visualization}

\begin{figure*}[]
	\centering
	\includegraphics[width=\linewidth]{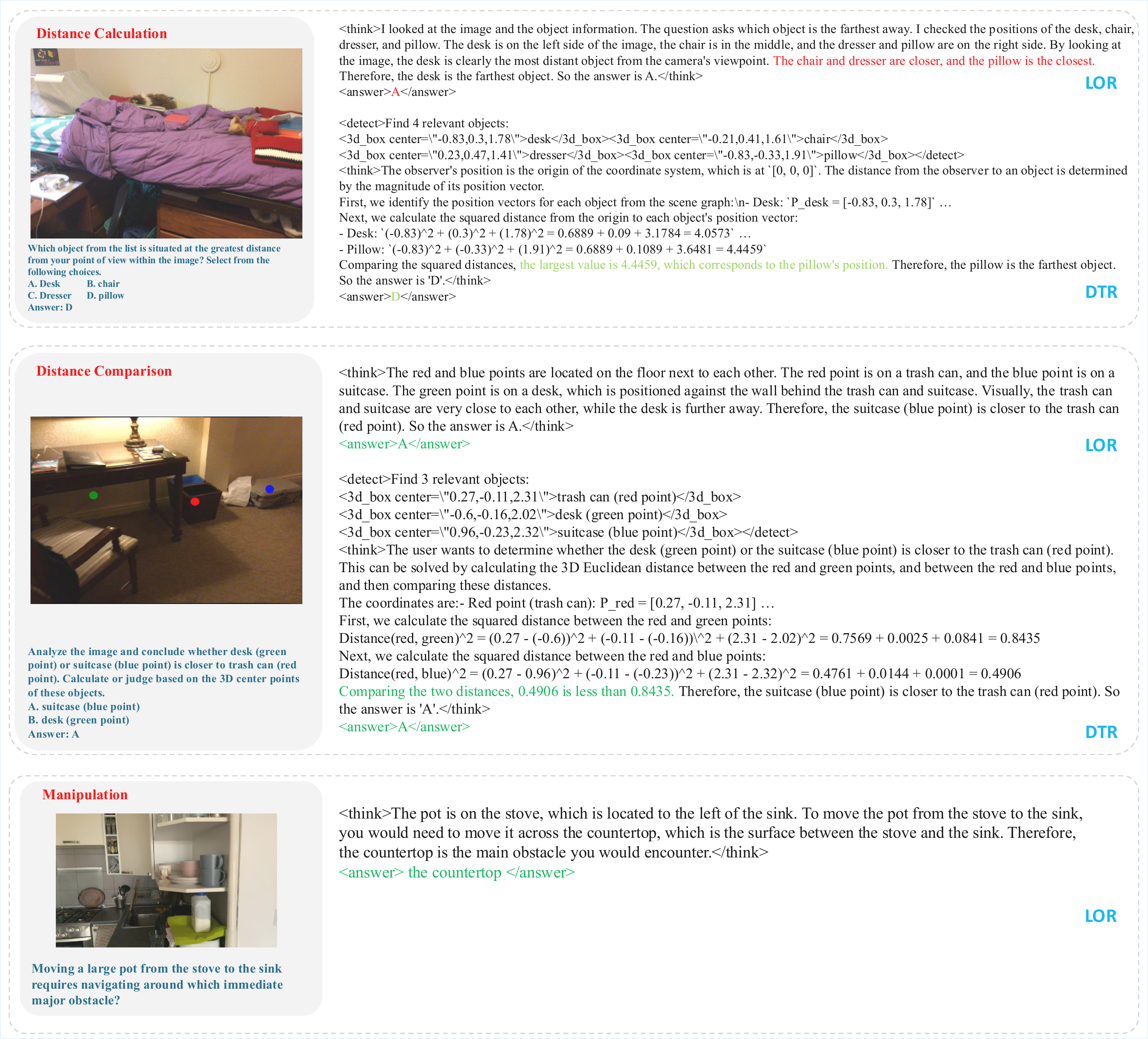}
	\caption{More visualization cases of SR-REAL.
    }
	\label{fig:appendix examples}
\end{figure*}

\textbf{Qualitative examples.} Figure \ref{fig:appendix examples} presents additional qualitative examples demonstrating the model's capability across both reasoning pathways. The model yields accurate results under both LOR and DTR paradigms. In the first example on distance measurement, LOR incorrectly estimates distance due to inaccurate object localization, while DTR accurately computes the distance through precise spatial coordinate detection. In the last example, which involves a manipulation task—specifically, identifying obstacles when moving a large pot from the stove to the sink—the model correctly infers that the manipulation trajectory intersects with the intermediate object (the countertop), thereby deriving the correct answer.

\noindent\textbf{Failure cases when inappropriate cold-start.} 
Tables \ref{tab:stage} and \ref{tab:cold start} in the main text demonstrate the critical impact of the cold-start phase on the model's final performance, highlighting that diverse cold-start data facilitates more stable reasoning on out-of-domain (OOD) questions. 
Table \ref{tab:fail cases} provides a detailed breakdown of failure modes. When RL is applied directly without a cold-start phase, the model frequently exhibits reasoning-answer inconsistency: it may select the correct option, yet the intermediate Chain-of-Thought (CoT) process is factually incorrect or logically misaligned with the answer. 
Furthermore, when the cold-start is restricted to CoT data based on SPAR, followed by RL on SPAR dataset, the model maintains accurate reasoning on in-domain queries but sometimes produces chaotic or incoherent CoT on OOD tasks. These findings underscore that incorporating a diverse mixture of data during both the cold-start and RL stages is essential for improving reasoning generalization.

\begin{table*}[ht]
\centering
\caption{Ablation of different mechanisms during LOR training. `Filter' indicates online filtering, and `KLcos' means the coefficient of the KL constraint term in GRPO undergoes cosine decay.}
\begin{adjustbox}{max width=\textwidth}
\begin{tabular}{ccccc}
\toprule
                & SPAR (SI) & SPAR (MI) & SAT   & EmbSpatial \\  \midrule
Cold-Start      & 56.53     & 45.89     & 62.67 & 65.66      \\
RL             & 69.88     & 66.96     & 64    & 76.2       \\ 
RL+filter       & \textbf{72.21}     & \textbf{69.39}     & 64.67 & 76.90      \\
RL+filter+KLcos & 71.48     & 68.35     & \textbf{66.67} & \textbf{77.66}  \\
\bottomrule
\end{tabular}
\end{adjustbox}
\label{tab: filter kl}
\end{table*}

\begin{table*}[ht]
\centering
\caption{Ablation of Grounding data in cold-start for DTR inference.}
\begin{adjustbox}{max width=\textwidth}
\begin{tabular}{cccc}
\toprule
                   & SPAR (SI) & EmbSpatial & CVBench (3D) \\
                   \midrule
Cold-Start         & \textbf{59.58}     & \textbf{68.54}      & \textbf{90.08}        \\
w/o Ground-data & 59.36     & 66.32      & 83.58    \\
\bottomrule
\end{tabular}
\end{adjustbox}
\label{tab: ground data}
\end{table*}

\begin{table*}[t]\centering
\caption{The system prompts for two reasoning paths of SR-REAL.}
\tiny
\begin{minipage}{1.0\textwidth}\vspace{0mm}    \centering

\begin{tcolorbox} 
    \centering
    \begin{tabular}{p{0.97\textwidth} c}

 \textbf{Language-Only Reasoning:}  & \\
You are a helpful assistant. The user asks a question, and then you solve it. \\
Please first think deeply about the question based on the given image, and then provide the final answer. The reasoning process and answer are enclosed within $<$think$>$ $<$/think$>$ and $<$answer$>$ $<$/answer$>$ tags, respectively, i.e., $<$think$>$ reasoning process here $<$/think$>$ $<$answer$>$ answer here $<$/answer$>$.\\
Question: \{Question\}
\\
\\
 \textbf{Detect-then-Reason:}  & \\

You are a helpful assistant. The user asks a question, and then you solve it. \\
Please first detect 3D centers of relevant objects, then think deeply about the question based on the given image, and finally provide the answer. The detection, reasoning and answer are enclosed within $<$detect$>$ $<$/detect$>$, $<$think$>$ $<$/think$>$ and $<$answer$>$ $<$/answer$>$ tags, respectively, i.e., $<$detect$>$ detection here $<$/detect$>$, $<$think$>$ reasoning process here $<$/think$>$ $<$answer$>$ answer here $<$/answer$>$.\\
Question: \{Question\}
    \end{tabular}
\end{tcolorbox}
\label{tab:sys prompt sr-real}
\end{minipage}
\end{table*}

\begin{table*}[t]\centering
\caption{The prompt for LVLM to produce LOR CoT.}
\tiny
\begin{minipage}{1.0\textwidth}\vspace{0mm}    \centering

\begin{tcolorbox} 
    \centering
    \begin{tabular}{p{0.97\textwidth} c}
\# ROLE: Spatial Reasoning Chain-of-thinking Generator \\

Given a multiple-choice spatial reasoning question, a scene image, and the correct answer option, generate a **step-by-step reasoning chain** that logically derives the correct choice. Your output must demonstrate precise spatial reasoning based on the provided scene context. \\
\\
\#\# INPUT CONTEXT: \\
You will receive the following:\\
1.  **Image:** An image depicting the 3D scene.\\
2.  **Question:** A multiple-choice question about the image.\\
3.  **Answer:** The ground truth answer index to the question. (e.g., "A", "B", "C", or "D")\\
\\
\#\# TASK:\\
Your primary task is to analyze 3D visual scenes within the image, understand the `Question` and `Answer` accurately, then generate chain-of-thinking for how to get the correct answer.\\
* Use the `Image` to understand scene context, object appearances, and spatial relationship.\\
* Use the `Answer` to validate your reasoning process, your reasoning must be able to arrive at the correct answer.\\
\\
\#\# GUIDELINES \& CONSTRAINTS: \\
* The reasoning process must be consistent with the spatial information in the image.\\
* The reasoning process must clearly solve the question, and derive the correct answer.\\
* The word number of reasoning process should be between 100 and 200.\\
* No need for explicit output of step 1, step 2 ...\\
\\
\#\# A REFERENCE FOR REASONING STRUCTURE\\
1. Scene interpretation\\
2. Analysis: Position (horizontal/vertical/depth), relationship, distance ...\\
3. Inference: solve the question, describe the thinking process\\
4. Conclusion: matching correct answer\\
\\
\#\# OUTPUT FORMAT:\\
Provide the output in the string format (the generated reaoning process).\\

    \end{tabular}
\end{tcolorbox}
\label{tab:gemini prompt LOR}
\end{minipage}
\end{table*}

\begin{table*}[t]\centering
\caption{The prompt for LVLM to produce DTR CoT.}
\tiny
\begin{minipage}{1.0\textwidth}\vspace{0mm}    \centering

\begin{tcolorbox} 
    \centering
    \begin{tabular}{p{0.97\textwidth} c}
\# ROLE: Spatial Reasoning Chain-of-thinking Generator\\

Given a multiple-choice spatial reasoning question, a scene image, the correct answer option, and center 3D coordinates of relevant objects, generate a **step-by-step reasoning chain** that logically derives the correct choice. Your output must demonstrate precise spatial reasoning based on the provided scene context and object positions.\\
\\
\#\# INPUT CONTEXT:\\
You will receive the following:\\
1.  **Image:** An image depicting the 3D scene.\\
2.  **Question:** A multiple-choice question about the image. The qustion is about some objects annotated by bounding boxes or points in the image.\\
3.  **Answer:** The ground truth answer index to the question. (e.g., "A", "B", "C", or "D")\\
4.  **Scene Graph:** A dictionary containing center 3D coordinates of object identifiers in the format: \{'red point': [-0.1, 0.2, 1.4]\}\\
\\
\#\# COORDINATE SYSTEM (Camera Coordinates):\\
* **+x axis:** Left to Right\\
* **+y axis:** Top to Bottom\\
* **+z axis:** Near to Far (depth)\\
* Note that the x y z coordinate is from the perspective of camera, if the question changes the observer to a different position and direction, the right, above, front of observer can not correspond with original x y z.\\
\\
\#\# TASK:\\
Your primary task is to analyze 3D visual scenes within the image, understand the `Question` and `Answer` accurately, combine center positions of objects, then generate chain-of-thinking for how to get the correct answer.\\
* Use the `Image` to understand scene context, object appearances, and spatial relationship.\\
* Use the `Scene Graph` to get precise 3D positions of objects mentioned in the question. The scene graph is about annotations (bbox or point), and you need to match the corresponding objects in the question.\\
* Use the `Answer` to validate your reasoning process, your reasoning must be able to arrive at the correct answer.\\
\\
\#\# GUIDELINES \& CONSTRAINTS: \\
* The reasoning process must be consistent with the spatial information in the image.\\
* The reasoning process must clearly solve the question, and derive the correct answer.\\
* The word number of reasoning process should be between 100 and 200.\\
* Keep the mathematical calculations brief, no need to output too many formulas (<3).\\
* Mainly use the visual content for relationship judgement, with 3D coordinate calculations as supplementary or for verification.\\
* No need for explicit output of step 1, step 2 ... or 1. 2. ...\\
\\
\#\# A REFERENCE FOR REASONING STRUCTURE\\
1. Scene interpretation\\
2. Analysis: Position (horizontal/vertical/depth), relationship, distance ...\\
3. Computaion: calculate or compare the 3D coordinates to support spatial reasoning\\
4. Inference: solve the question, describe the thinking process\\
5. Conclusion: matching correct answer, e.g., "So the answer is 'A'."\\
\\
\#\# OUTPUT FORMAT:\\
Provide the reasoning process in the string format:
* (String) The generated explanation of how the answer was derived, referencing image context and 3D coordinates of objects. The word number of reasoning process should be less than 200.

    \end{tabular}
\end{tcolorbox}
\label{tab:gemini prompt DTR}
\end{minipage}
\end{table*}

\begin{table*}[t]\centering
\caption{The prompt for LVLM to generate complex spatial task questions.}
\tiny 
\begin{minipage}{1.0\textwidth}\vspace{0mm}    \centering
\begin{tcolorbox} 
    \centering
    \begin{tabular}{p{0.97\textwidth} c}
\# ROLE: Spatial Reasoning Evaluator Question Generator \\

\#\# PERSONA:\\
You are an expert evaluator specializing in designing challenging spatial reasoning tasks for advanced AI models. You are creative, meticulous, and focused on probing deep spatial understanding beyond simple object recognition.
\\
\#\# INPUT:\\
You will receive an image or a video.\\

\#\# TASK:\\
Generate exactly 6 highly challenging spatial reasoning questions based on the provided image or video. These questions are intended to evaluate the capabilities of a state-of-the-art vision-language model. The questions must necessitate multi-step reasoning, inference about implicit spatial relationships, object interactions, potential future states, or navigation planning within the scene.\\

\#\# QUESTION CATEGORIES \& DISTRIBUTION:\\
Generate questions covering the following categories. Unless a specific type is requested by the user later, adhere to this distribution:\\
* **Navigation (2 questions):** Focus on planning paths, anticipating obstacles, understanding traversable space, or correcting navigational errors based on visual cues.\\
    * *Example Concept:* "Outline a path from [Object A] to [Object B], staying only on [Surface Type] and going around [Obstacle]."\\
    * *Example Concept:* "If attempting to move from [Location X] to [Location Y] following [Simple Rule], what is the first major obstacle encountered?"\\
* **Manipulation (2 questions):** Focus on object affordances, interaction plausibility, planning sequences of actions for rearrangement, or predicting outcomes of physical interactions.\\
    * *Example Concept:* "Could the [Object C] realistically be picked up with one hand and placed securely on top of [Object D]?"\\
    * *Example Concept:* "Describe the steps to rearrange [Objects E, F, G] from their current positions to form a [Target Configuration]."\\
* **Object Search \& Planning (1 question):** Focus on locating objects (potentially occluded or partially visible) and planning how to access them.\\
    * *Example Concept:* "Where is the most likely place to find a [Common Object Type] in this scene, and is it currently visible?"\\
* **Scene Arrangement/Layout Reasoning (1 question):** Focus on the plausibility of object placements, spatial relationships between objects, or understanding the functional layout of the scene.\\
    * *Example Concept:* "Is the placement of [Object H] relative to [Object I] typical for this type of environment? Why or why not?"\\
    * *Example Concept:* "If you added a [New Object], where could it fit without disturbing the main function of the area near [Existing Object J]?"\\
\#\# CONSTRAINTS:\\
1.  **Deep Reasoning Required:** Questions MUST demand more than simple identification or direct observation. They should require inferring relationships, predicting outcomes, or planning sequences.\\
2.  **Visual Grounding Essential:** Questions MUST require information *directly* from the image/video. They cannot be answerable using only common sense or external knowledge *without* visual context.\\
3.  **Integration of Knowledge:** Questions SHOULD ideally require the model to integrate visual information with general world knowledge (e.g., physics, object properties, typical layouts) to arrive at the answer.\\
4.  **3D Spatial Computation Necessary:** Questions MUST require reasoning about 3D spatial attributes (coordinates, dimensions, orientation) and involve numerical calculations (distances, angles, volumetric comparisons) derived from the visual data.\\
5.  **Minimal Explicit Information:** Phrase questions concisely, providing only necessary context. Avoid giving away information that the evaluated model should infer from the visual input.\\
6.  **Conciseness:** Each question must be a maximum of 25 words. (Slightly increased from 20 for flexibility).\\
7.  **Creativity:** Strive for novel and non-obvious questions that genuinely test advanced spatial understanding.\\

\#\# INTERNAL PROCESS GUIDELINE:\\
* To aid in generating complex questions, you may first internally generate a detailed description or caption of the image/video, focusing on objects, spatial relationships, potential interactions, and environmental context. Do NOT include this caption in your final output.\\

\#\# OUTPUT FORMAT:\\
* Provide the 6 questions as a numbered list.\\
* Indicate the category for each question in parentheses, e.g., `1. Question text (Navigation)`

    \end{tabular}
\end{tcolorbox}
\label{tab:sys prompt questions}
\end{minipage}
\end{table*}

\begin{table*}[t]\centering
\caption{The failure cases when inappropriate cold-start.}
\tiny
\begin{minipage}{1.0\textwidth}\vspace{0mm}    \centering

\begin{tcolorbox} 
    \centering
    \begin{tabular}{p{0.97\textwidth} c}

 \textbf{Failure cases when RL only:}  & \\
Question: What changes in the observer’s perception of the spatial relationship between monitor (red bbox) and pillow (yellow bbox) when observer relocates to pillow (green bbox) and face window (blue bbox)? Calculate or judge based on the 3D center points of these objects. Please use the world coordinate system to determine the up and down position relationship of objects.
Pick the right response from the available choices.\\
A. right, above, closer\\
B. right, , farther\\
C. left, above, \\
D. , above, closer\\
\\
Response: $<$think$>$The observer's perspective changes as they move from the pillow to the window. Initially, the monitor is to the right of the pillow and below it. When the observer moves to the window, the monitor is now below the observer's current position, making it closer to the observer. Therefore, the observer's perception of the spatial relationship between the monitor and the pillow changes to \textcolor{red}{right, above, closer}.$<$/think$>$ $<$answer$>$\textcolor{green}{B}$<$/answer$>$
\\
\\
 \textbf{Failure cases in out-of-domain questions when cold-start with only SPAR cot:}  & \\
question: Which motion direction should the spoon be moved in order to slice some cheese? Choices:  A. orange. B. red. C. blue. D. light blue. Please answer directly with only the letter of the correct option and nothing else.\\
\\
Response: $<$think$>$The red arrow points towards the lid of the container, indicating the direction of$<$think$>$The red arrow points towards the lid, suggesting the direction of$<$think$>$The red arrow$<$th$<$th$<$th$<$th$<$th$<$th$<$th$<$th$<$th$<$th$<$th$<$th$<$th$<$th$<$th$<$th$<$th$<$th$<$th$<$th$<$th$<$th$<$th$<$th$<$th$<$th\\$<$th$<$th$<$th$<$th$<$th$<$th$<$th$<$th$<$th$<$th$<$th$<$th$<$th$<$th$<$th$<$th$<$th$<$th$<$th$<$th$<$th$<$th$<$th$<$th$<$th$<$th$<$th

    \end{tabular}
\end{tcolorbox}
\label{tab:fail cases}
\end{minipage}
\end{table*}

\section{Limitations}
\label{app:limitation}

While \method{} demonstrates strong spatial reasoning performance, several limitations remain.
First, our framework is built upon the SR-3D spatial VLM, which requires depth maps and camera intrinsics/extrinsics at inference time. This limits applicability to settings where such geometric metadata is unavailable, such as unconstrained in-the-wild images without depth sensors.
Second, the DTR path relies on region tokens derived from 2D bounding box or mask annotations. When such region annotations are absent—as in global benchmarks like SAT—DTR cannot be applied, and the model falls back to LOR. Automatically generating reliable region proposals for arbitrary queries remains an open challenge.
Third, despite gains on OOD benchmarks, we observe that applying chain-of-thought reasoning to perception-heavy tasks (e.g., BLINK, RealWorldQA) can hurt performance compared to direct inference. This suggests the model has not fully learned when to engage multi-step reasoning versus respond directly, an important direction for future work.
Finally, the cold-start data construction relies on a proprietary LVLM (Gemini-2.5-Pro) for CoT generation, introducing a dependency on external APIs and potential quality variance across reasoning traces.

\section{Broader Impact}
\label{app:impact}

\noindent\textbf{Positive impacts.}
Improving spatial reasoning in vision-language models has broad beneficial applications. Enhanced spatial understanding supports embodied AI systems, assistive robotics, autonomous driving, and augmented reality, all of which can improve accessibility and quality of life. More capable spatial reasoning may also accelerate progress in scientific domains that require interpreting 3D spatial data, such as medical imaging analysis and remote sensing.

\noindent\textbf{Potential negative impacts.}
As with general-purpose vision-language models, improvements in spatial understanding could be misused in surveillance systems or automated physical-space monitoring without appropriate consent. We encourage responsible deployment and advocate for clear usage guidelines when applying spatially-aware models in sensitive or privacy-sensitive environments. The reliance on proprietary data generation pipelines may also concentrate capability development among well-resourced organizations, potentially widening access gaps.